\documentclass[runningheads]{llncs}
\usepackage{graphicx}
\usepackage{subcaption}
\usepackage{algorithm} 
\usepackage{algorithmicx}
\usepackage{algpseudocode}
\usepackage[utf8]{inputenc}
\usepackage{csquotes}
\usepackage{amsmath}
\usepackage{amsfonts}
\usepackage{comment}
\usepackage{mdframed}
\usepackage{mathtools}
\usepackage{braket}
\usepackage{varwidth}
\usepackage{placeins}
\newcommand\numberthis{\addtocounter{equation}{1}\tag{\theequation}}

\begin{document}
\title{Metatrace Actor-Critic: Online Step-size Tuning\\by Meta-gradient Descent\\for Reinforcement Learning Control}
\titlerunning{Metatrace}
\author{Kenny Young, %\orcidID{0000-1111-2222-3333},
Baoxiang Wang, \and %\orcidID{0000-1111-2222-3333} \and
Matthew E.~Taylor} %\orcidID{1111-2222-3333-4444}}
\institute{
Borealis AI, Edmonton, Alberta, Canada \\
\email{\{kenny.young, brandon.wang, matthew.taylor\}@BorealisAI.com}}
\maketitle              % typeset the header of the contribution
%
%A variety of optimization methods have been derived for training neural networks with stochastic gradient descent for the supervised learning problem. These are often applied to reinforcement learning problems in a black box fashion, but often the same assumptions do not hold. Much of the recent Deep RL literature involves making the reinforcement learning problem as close as possible to the supervised learning problem so that similar methods are applicable. In contrast we derive a set of step-size tuning algorithms specifically for online reinforcement learning with eligibility traces, based on meta-gradient descent. The technique, Metatrace, makes use of an eligibility trace analogous to that used in methods like $TD(\lambda)$. We explore tuning both a single scalar step-size and a separate step-size for each network weight. We evaluate Metatrace first for control with linear function approximation in the classic mountain car problem and a noisy-nonstationary version. Finally, we apply Metatrace for control with nonlinear function approximation in 5 games in the Arcade Learning Environment and explore how it impacts learning speed, and robustness to initial step-size choice.

\begin{abstract}
% Reinforcement learning (RL) has had many successes in both ``shallow'' and ``deep'' settings. Unfortunately, in both cases, there are many parameters that must be tuned so that the agents can achieve reasonable performance. While there are many rules of thumb for how to set these parameters, high-quality results typically require large parameter tuning searches. To attempt to combat such extensive parameter searches, this work focuses on applying meta-learning techniques to RL settings. In particular, we derive a set of step-size tuning algorithms specifically for online RL with eligibility traces. Our novel technique, Metatrace, makes use of an eligibility trace analogous to that used in methods like $TD(\lambda)$. We explore tuning both a single scalar step-size and a separate step-size for each learned parameter. We evaluate Metatrace first for control with linear function approximation in the classic mountain car problem and a noisy-nonstationary version. Finally, we apply Metatrace for control with nonlinear function approximation in 5 games in the Arcade Learning Environment and explore how it impacts learning speed, and robustness to initial step-size choice. Results show that the Metatrace parameter is relatively easy to set and \MET{...}
Reinforcement learning (RL) has had many successes in both ``deep'' and ``shallow'' settings. In both cases, significant hyperparameter tuning is often required to achieve good performance. Furthermore, when nonlinear function approximation is used, non-stationarity in the state representation can lead to learning instability. A variety of techniques exist to combat this --- most notably large experience replay buffers or the use of multiple parallel actors. These techniques come at the cost of moving away from the online RL problem as it is traditionally formulated (i.e., a single agent learning online without maintaining a large database of training examples). Meta-learning can potentially help with both these issues by tuning hyperparameters online and allowing the algorithm to more robustly adjust to non-stationarity in a problem. This paper applies meta-gradient descent to derive a set of step-size tuning algorithms specifically for online RL control with eligibility traces. Our novel technique, Metatrace, makes use of an eligibility trace analogous to methods like $TD(\lambda)$. We explore tuning both a single scalar step-size and a separate step-size for each learned parameter. We evaluate Metatrace first for control with linear function approximation in the classic mountain car problem and then in a noisy, non-stationary version. Finally, we apply Metatrace for control with nonlinear function approximation in 5 games in the Arcade Learning Environment where we explore how it impacts learning speed and robustness to initial step-size choice. Results show that the meta-step-size parameter of Metatrace is easy to set, Metatrace can speed learning, and Metatrace can allow an RL algorithm to deal with non-stationarity in the learning task.

\end{abstract}
\setlength{\textfloatsep}{10pt}
\section{Introduction}
In the supervised learning (SL) setting, there are a variety of optimization methods that build on stochastic gradient descent (SGD) for tuning neural network (NN) parameters (e.g., RMSProp \cite{RMSProp} and ADAM \cite{ADAM}). These methods generally aim to accelerate learning by monitoring gradients and modifying updates such that the effective loss surface has more favorable properties.

Most such methods are derived for SGD on a fixed objective (i.e., average loss over a training set). This does not translate directly to the online reinforcement learning (RL) problem, where targets incorporate future estimates, and subsequent observations are correlated. Eligibility traces complicated this further, as individual updates no longer correspond to a gradient descent step toward any target on their own. Eligibility traces break up the target into a series of updates such that only the sum of updates over time moves toward it.

To apply standard SGD techniques in the RL setting, a common strategy is to make the RL problem as close to the SL problem as possible. Techniques that help achieve this include: multiple actors \cite{A3C}, large experience replay buffers \cite{DQN}, and separate online and target networks \cite{DDQN}. These all help smooth gradient noise and mitigate non-stationarity such that SL techniques work well. They are not, however, applicable to the more standard RL setting where a single agent learns online without maintaining a large database of training examples.

This paper applies meta-gradient descent, propagating gradients through the optimization algorithm itself, to derive step-size tuning algorithms specifically for the RL control problem. We derive algorithms for this purpose based on the IDBD approach \cite{idbd}. We refer to the resulting methods as Metatrace algorithms.

Using this novel approach to meta-gradient descent for RL control we define algorithms for tuning a scalar step-size, as well as a vector of step-sizes (one element for each parameter), and finally a mixed version which aims to leverage the benefits of both. Aside from these algorithms, our main contributions include applying meta-gradient descent to actor-critic with eligibility traces (\textit{AC}$(\lambda)$), and exploring the performance of meta-gradient descent for RL with a non-stationary state representation, including with nonlinear function approximation (NLFA). In particular, we evaluate Metatrace with linear function approximation (LFA) for control in the classic mountain car problem and a noisy, non-stationary variant. We also evaluate Metatrace for training a deep NN online in the 5 original training set games in the Arcade Learning Environment (ALE) \cite{ALE}, with eligibility traces and without using either multiple actors or experience replay.

\section{Related Work}
Our work is closely related to IDBD \cite{idbd} and its extension autostep \cite{autostep}, meta-gradient decent procedures for step-size tuning in the supervised learning case. Even more closely related, are SID and NOSID \cite{dabney}, analogous meta-gradient descent procedures for \textit{SARSA}$(\lambda)$. Our approach differs primarily by explicitly accounting for time-varying weights in the optimization objective for the step-size. In addition, we extend the approach to \textit{AC}$(\lambda)$ and to vector-valued step-sizes as well as a ``mixed'' version which utilizes a combination of scalar and vector step-sizes. Also related are TIDBD and it's extension AutoTIDBD \cite{TIDBD,TIDBD2}, to our knowledge the only prior work to investigate learning of vector step-sizes for RL. The authors focuses on \textit{TD}$(\lambda)$ for prediction, and explore both vector and scalar step-sizes. They demonstrate that for a broad range of parameter settings, both scalar and vector AutoTIDBD outperform ordinary TD$(\lambda)$, while vector AutoTIDBD outperforms a variety of scalar step-size adaptation methods and TD$(\lambda)$ with an optimal fixed step-size. Aside from focusing on control rather than prediction, our methods differs from TIDBD primarily in the objective optimized by the step-size tuning. They use one step TD error; we use a multistep objective closer to that used in SID. Another notable algorithm, crossprop \cite{crossprop}, applies meta-gradient descent for directly learning good features from input\footnote{Crossprop is used in place of backprop to train a single hidden layer.}, as opposed to associated step-sizes. The authors demonstrate that using crossprop in place of backprop can result in feature representations which are more robust to non-stationarity in the task. Our NN experiments draw inspiration from \cite{SiLU}, to our knowledge the only prior work to apply online RL with eligibility traces\footnote{In their case, they use \textit{SARSA}$(\lambda)$.} to train a modern deep NN.

\section{Background}\label{background}
We consider the RL problem where a learning agent interacts with an environment while striving to maximize a reward signal. The problem is generally formalized as a Markov Decision Process described by a 5-tuple: $\left<\mathcal{S},\mathcal{A},p,r,\gamma\right>$. At each time-step the agent observes the state $S_t\in\mathcal{S}$ and selects an action $A_t\in \mathcal{A}$. Based on $S_t$ and $A_t$, the next state $S_{t+1}$ is generated, according to a probability $p(S_{t+1}|S_{t},A_{t})$. The agent additionally observes a reward $R_{t+1}$, generated by $r:\mathcal{S}\times\mathcal{A}\rightarrow \mathbb{R}$. Algorithms for reinforcement learning broadly fall into two categories, prediction and control. In prediction the agent follows a fixed policy $\pi:\mathcal{S}\times\mathcal{A}\rightarrow [0,1]$ and seeks to estimate from experience the expectation value of the return $G_t=\sum\limits_{k=t}^{\infty}\gamma^{k-t}R_{k+1}$, with discount factor $\gamma\in[0,1]$. In control, the goal is to learn, through interaction with the initially unknown environment, a policy $\pi$ that maximizes the expected return $G_t$, with discount factor $\gamma\in[0,1]$. In this work we will derive step-size tuning algorithms for the control case.

Action-value methods like Q-learning are often used for RL control. However, for a variety of reasons, actor-critic (AC) methods are becoming increasingly more popular in Deep RL --- we will focus on AC. AC methods separately learn a state value function for the current policy and a policy which attempts to maximize that value function. In particular, we will derive Metatrace for actor critic with eligibility traces, \textit{AC}$(\lambda)$ \cite{AC_lambda,GAE}. While eligibility traces are often associated with prediction methods like \textit{TD}$(\lambda)$ they are also applicable to AC.

To specify the objective of \textit{TD}$(\lambda)$, and by extension \textit{AC}$(\lambda)$, we must first define the lambda return $G_{w,t}^{\lambda}$. Here we will define $G_{w,t}^{\lambda}$ associated with a particular set of weights $w$ recursively:
\begin{equation*}
G_{w,t}^{\lambda}=R_{t+1}+\gamma\left((1-\lambda)V_w(S_{t+1}) +\lambda G^{\lambda}_{w,t+1}\right)
\end{equation*}
$G_{w,t}^{\lambda}$ bootstraps future evaluations to a degree controlled by $\lambda$.  If $\lambda < 1$, then $G_{w,t}^{\lambda}$ is a biased estimate of the return, $G_t$.  If $\lambda = 1$, then $G_{w,t}^{\lambda}$ reduces to $G_t$. Here we define $G_{w,t}^{\lambda}$ for a fixed weight $w$; in Section \ref{alg} we will extend this to a time varying $w_t$. Defining TD-error, $\delta_t = R_t+\gamma V_w(S_{t+1})-V_w(S_t)$, we can expand $G_{w,t}^{\lambda}$ as the current state value estimate plus the sum of future discounted $\delta_t$ values:
\begin{equation*}
G_{w,t}^{\lambda}=V_w(S_t)+\sum_{k=t}^{\infty}(\gamma\lambda)^{k-t}\delta_k
\end{equation*}
This form is useful in the derivation of \textit{TD}$(\lambda)$ as well as \textit{AC}$(\lambda)$. \textit{TD}$(\lambda)$ can be understood as minimizing the mean squared error $\left(G_{w,t}^{\lambda}-V_w(S_t)\right)^2$ between the value function $V_{w}$ (a function of the current state $S_t$ parameterized by weights $w$) and the lambda return $G_{w,t}^{\lambda}$. In deriving \textit{TD}$(\lambda)$, the target $G_{w,t}^{\lambda}$ is taken as constant despite its dependence on $w$. For this reason, \textit{TD}$(\lambda)$ is often called a ``semi-gradient'' method. Intuitively, we want to modify our current estimate to match our future estimates and not the other way around. For \textit{AC}$(\lambda)$, we will combine this mean squared error objective with a policy improvement term, such that the combined objective represents a trade-off between the quality of our value estimates and the performance of our policy:
\begin{equation}\label{control_objective}
\mathcal{J}_{\lambda}(w)=\frac{1}{2}\left(\sum\limits_{t=0}^{\infty}\left(G_{w,t}^{\lambda}-V_w(S_t)\right)^2-\sum\limits_{t=0}^{\infty}\log(\pi_w\left(A_t\middle|S_t\right))\left(G_{w,t}^{\lambda}-V_w(S_t)\right)\right)
\end{equation}
As in \textit{TD}$(\lambda)$, we apply the notion of a semi-gradient to optimizing equation \ref{control_objective}. In this case along with $G_{w,t}^{\lambda}$, the appearance of $V_{w}(S_t)$ in the right sum is taken to be constant. Intuitively, we wish to improve our actor under the evaluation of our critic, not modify our critic to make our actor's performance look better. With this caveat in mind, by the policy gradient theorem \cite{policy_grad}, the expectation of the gradient of the right term in equation \ref{control_objective} is approximately equal to the (negated) gradient of the expected return. This approximation is accurate to the extent that our advantage estimate $\left(G_{w,t}^{\lambda}-V_{w}(S_t)\right)$ is accurate. Descending the gradient of the right half of $\mathcal{J}_{\lambda}(w)$ is then ascending the gradient of an estimate of expected return. Taking the semi-gradient of equation \ref{control_objective} yields:
\begin{align*}
\frac{\partial}{\partial w}\mathcal{J}_{\lambda}(w)
&=-\sum\limits_{t=0}^{\infty}\left(\frac{\partial V_{w}(S_t)}{\partial w}+\frac{1}{2}\frac{\partial \log(\pi_{w}\left(A_t\middle|S_t\right))}{\partial w}\right)\left(G_{t}^{\lambda}-V_{w}(S_t)\right)\\
&=-\sum\limits_{t=0}^{\infty}\left(\frac{\partial V_{w}(S_t)}{\partial w}+\frac{1}{2}\frac{\partial\log(\pi_{w}\left(A_t\middle|S_t\right))}{\partial w}\right)\sum\limits_{k=t}^{\infty}(\gamma\lambda)^{k-t}\delta_k\\
&=-\sum\limits_{t=0}^{\infty}\delta_t\sum\limits_{k=0}^{t}(\gamma\lambda)^{t-k}\left(\frac{\partial V_{w}(S_t)}{\partial w}+\frac{1}{2}\frac{\partial\log(\pi_{w}\left(A_t\middle|S_t\right))}{\partial w}\right)
\end{align*}
Now define for compactness $U_w(S_t)\dot{=}V_w(S_t)+\frac{1}{2}\log(\pi_{w}\left(A_t\middle|S_t\right))$ and define the eligibility trace at time $t$ as $z_t=\sum\limits_{k=0}^{t}(\gamma\lambda)^{t-k} \frac{\partial U_w(S_k)}{\partial w}$, such that:
\begin{equation}
\label{TD_grad}
\frac{\partial}{\partial w}\mathcal{J}_{\lambda}(w)=-\sum\limits_{t=0}^{\infty}\delta_t z_t
\end{equation}
Offline \textit{AC}$(\lambda)$ can be understood as performing a gradient descent step along equation \ref{TD_grad}. Online \textit{AC}$(\lambda)$, analogous to online \textit{TD}$(\lambda)$ can be seen as an approximation to this offline version (exact in the limit of quasi-static weights) that updates weights after every time-step. Advantages of the online version include making immediate use of new information, and being applicable to continual learning\footnote{Where an agent interacts with an environment indefinitely, with no distinct episodes.}. Online \textit{AC}$(\lambda)$ is defined by the following set of equations:
\begin{align*}
z_{t}&=\gamma\lambda z_{t-1}+\frac{\partial U_{w_t}(S_t)}{\partial w_t}\\
w_{t+1}&=w_t+\alpha z_{t}\delta_t
\end{align*}

\section{Algorithm}
\label{alg}
We will present three variations of Metatrace for control using \textit{AC}$(\lambda)$, scalar (single $\alpha$ for all model weights), vector (one $\alpha$ per model weight), and finally a ``mixed'' version that attempts to leverage the benefits of both. Additionally, we will discuss two practical improvements over the basic algorithm, normalization that helps to mitigate parameter sensitivity across problems and avoid divergence, and entropy regularization which is commonly employed in actor-critic to avoid premature convergence \cite{A3C}.

\subsection{Scalar Metatrace for \textit{AC}$(\lambda)$} Following \cite{idbd}, we define our step-size as $\alpha=e^{\beta}$. For tuning $\alpha$, it no longer makes sense to define our objective with respect to a fixed weight vector $w$, as in equation \ref{control_objective}. We want to optimize $\alpha$ to allow our weights to efficiently track the non-stationary \textit{AC}$(\lambda)$ objective. To this end we define the following objective incorporating time-dependent weights:\vspace{-3pt}
\begin{align*}
\mathcal{J}^{\beta}_{\lambda}&(w_0..w_{\infty})\\
&=\frac{1}{2}\left(\sum\limits_{t=0}^{\infty}\left(G_{t}^{\lambda}-V_{w_t}(S_t)\right)^2-\sum\limits_{t=0}^{\infty}\log(\pi_{w_t}\left(A_t\middle|S_t\right))\left(G_{t}^{\lambda}-V_{w_t}(S_t)\right)\right)\numberthis\label{meta_objective}
\end{align*}
Here, $G_t^{\lambda}$ with no subscript $w$ is defined as $G_t^{\lambda}=V_{w_t}(S_t)+\sum_{k=t}^{\infty}(\gamma\lambda)^{k-t}\delta_k$, where $\delta_k=R_k+\gamma V_{w_k}(S_{k+1})-V_{w_k}(S_k)$. We will follow a derivation similar to Section 4.3.1 of \cite{dabney}, but the derivation there does not explicitly account for the time dependence of $w_t$. Instead, in equation 4.18 they differentiate $\mathcal{J}_{\lambda}(w_t)$ with respect to $\alpha$ as follows\footnote{Bra-ket notation ($\left<\cdot\middle|\cdot\right>$) indicates dot product.}:\vspace{-3pt}
\begin{align*}
\frac{\partial}{\partial \alpha} \mathcal{J}_{\lambda}(w_t)&=\left<\frac{\partial}{\partial w_t}\mathcal{J}_{\lambda}(w_t)\middle|\frac{\partial w_t}{\partial \alpha}\right>\\
&=-\sum\limits_{t=0}^\infty\left(\delta_t\left<z_t\middle|\frac{\partial w_t}{\partial \alpha}\right>\right)
\end{align*}
The first line applies the chain rule with $\mathcal{J}_{\lambda}(w_t)$ treated as a function of a single $w_t$ vector. The second line is unclear, in that it takes $\frac{\partial w_t}{\partial \alpha}$ inside the sum in equation \ref{TD_grad}. The time index of the sum is not \textit{a priori} the same as that of $w_t$. We suggest this ambiguity stems from propagating gradients through an objective defined by a single $w_t$ value, while the significance of $\alpha$ is that it varies the weights over time. For this reason, we hold that it makes more sense to minimize equation \ref{meta_objective} to tune $\alpha$. We will see in what follows that this approach yields an algorithm very similar to \textit{TD}$(\lambda)$ for tuning the associated step-sizes.

Consider each $w_t$ a function of $\beta$, differentiating equation \ref{meta_objective}, with the same semi-gradient treatment used in \textit{AC}$(\lambda)$ yields:\vspace{-3pt}
\begin{align*}
\frac{\partial}{\partial \beta}\mathcal{J}_{\lambda}(w_0..w_{\infty})
&=-\sum\limits_{t=0}^{\infty}\left(\frac{\partial V_{w_t}(S_t)}{\partial\beta}+\frac{1}{2}\frac{\log(\pi_{w_t}\left(A_t\middle|S_t\right))}{\partial\beta}\right)\left(G_{t}^{\lambda}-V_{w_t}(S_t)\right)\\
&=-\sum\limits_{t=0}^{\infty}\left<\frac{\partial U_{w_t}(S_t)}{\partial w_t}\middle|\frac{\partial w_t}{\partial \beta}\right>\sum\limits_{k=t}^{\infty}(\gamma\lambda)^{k-t}\delta_k\\
&=-\sum\limits_{t=0}^{\infty}\delta_t\sum\limits_{k=0}^{t}(\gamma\lambda)^{t-k}\left<\frac{\partial U_{w_t}(S_t)}{\partial w_t}\middle|\frac{\partial w_t}{\partial \beta}\right>
\end{align*}
Now, define a new eligibility trace. This time as:\vspace{-3pt}
\begin{equation}\label{metatrace}
z_{\beta,t}\dot{=}\sum\limits_{k=0}^{t}(\gamma\lambda)^{t-k}\left<\frac{\partial U_{w_k}(S_k)}{\partial w_k}\middle|\frac{\partial w_k}{\partial \beta}\right>
\end{equation}\vspace{-2pt}
such that:\vspace{-3pt}
\begin{equation}\label{control_grad}
\frac{\partial}{\partial \beta}\mathcal{J}_{\lambda}(w_0..w_{\infty})=-\sum\limits_{t=0}^{\infty}\delta_tz_{\beta,t}
\end{equation}\vspace{-2pt}

To compute $h(t)\dot{=}\frac{\partial w_t}{\partial \beta}$, we use a method analogous to that in \cite{dabney}. As $z$ itself is a sum of first order derivatives with respect to $w$, we will ignore the effects of higher order derivatives and approximate $\frac{\partial z_t}{\partial w}=0$. Since $\log(\pi_{w_t}\left(A_t\middle|S_t\right))$ necessarily involves some non-linearity, this is only a first order approximation even in the LFA case. Furthermore, in the control case, the weights affect action selection. Action selection in turn affects expected weight updates. Hence, there are additional higher order effects of modifying $\beta$ on the expected weight updates, and like \cite{dabney} we do not account for these effects. We leave open the question of how to account for this interaction in the online setting, and to what extent it would make a difference. We compute $h(t)$ as follows:\vspace{-3pt}
\begin{align*}
h(t+1)&=\frac{\partial}{\partial \beta}[w_t+\alpha\delta_t z_t]\\
&\approx h(t)+\alpha\delta_t z_t+\alpha z_t \frac{\partial\delta_t}{\partial \beta}\\
&=h(t)+\alpha z_t\left(\delta_t +\left<\frac{\partial \delta_t}{\partial w_t}\middle|\frac{\partial w_t}{\partial \beta}\right>\right)\\
&=h(t)+\alpha z_t\left(\delta_t +\left<\gamma\frac{\partial V_{w_t}(S_{t+1})}{\partial w_t}-\frac{\partial V_{w_t}(S_{t})}{\partial w_t}\middle|h(t)\right>\right)\numberthis\label{scalar_h}\\
\end{align*}\vspace{-3pt}

Note that here we use the full gradient of $\delta_t$ as opposed to the semi-gradient used in computing equation \ref{control_grad}. This can be understood by noting that in equation \ref{control_grad} we effectively treat $\alpha$ similarly to an ordinary parameter of $AC(\lambda)$, following a semi-gradient for similar reasons\footnote{As explained in Section \ref{background}.}. $h(t)$ on the other hand is meant to track, as closely as possible, the actual impact of modifying $\alpha$ on $w$, hence we use the full gradient and not the semi-gradient. All together, Scalar Metatrace for \textit{AC}$(\lambda)$ is described by:\vspace{-3pt}
\begin{align*}
z_{\beta}&\leftarrow \gamma\lambda z_{\beta}+\left<\frac{\partial U_{w_t}(S_t)}{\partial w_t}\middle|h\right>\\
\beta&\leftarrow \beta+\mu z_{\beta}\delta_t\\
h&\leftarrow h+e^{\beta}z\left(\delta_t +\left<\gamma\frac{\partial V_{w_t}(S_{t+1})}{\partial w_t}-\frac{\partial V_{w_t}(S_{t})}{\partial w_t}\middle|h\right>\right)
\end{align*}
The update to $z_{\beta}$ is an online computation of equation \ref{metatrace}. The update to $\beta$ is exactly analogous to the \textit{AC}$(\lambda)$ weight update but with equation \ref{control_grad} in place of equation \ref{TD_grad}. The update to $h$ computes equation \ref{scalar_h} online. We will augment this basic algorithm with two practical improvements, entropy regularization and normalization, that we will discuss in the following subsections.

\subsubsection{Entropy Regularization}\label{entropy_reg}
In practice it is often helpful to add an entropy bonus to the objective function to discourage premature convergence \cite{A3C}. Here we cover how to modify the step-size tuning algorithm to cover this case. We observed that accounting for the entropy bonus in the meta-optimization (as opposed to only the underlying objective) improved performance in the ALE domain. Adding an entropy bonus with weight $\psi$ to the actor critic objective of equation 2 gives us:\vspace{-3pt}
\begin{equation}\label{control_objective_entropy}
\mathcal{J}_{\lambda}(w)=\begin{multlined}[t]\frac{1}{2}\left(\sum\limits_{t=0}^{\infty}\left(G_{t}^{\lambda}-V_{w}(s_t)\right)^2-\sum\limits_{t=0}^{\infty}\log(\pi_{w}\left(A_t\middle|S_t\right))\left(G_{t}^{\lambda}-V_{w}(S_t)\right)\right)\\
-\psi \sum\limits_{t=0}^{\infty}H_{w}(S_t)
\end{multlined}
\end{equation}
with $H_{w_t}(S_t)=-\sum\limits_{a\in\mathcal{A}} \pi_{w_t}\left(a\middle|S_t\right)\log(\pi_{w_t}\left(a\middle|S_t\right))$.
The associated parameter update algorithm becomes:\vspace{-3pt}
\begin{align*}
z&\leftarrow\gamma\lambda z+\frac{\partial U_{w_t}(s_t)}{\partial w_t}\\
w&\leftarrow w+\alpha\left(z\delta_t+\psi \frac{\partial H_{w_t}(s_t)}{\partial w_t}\right)
\end{align*}
To modify the meta update algorithms for this case, modify equation \ref{control_objective_entropy} with time dependent weights:\vspace{-3pt}
\begin{align*}
\mathcal{J}^{\beta}_{\lambda}&(w_0..w_{\infty})\\
&=\begin{multlined}[t]\frac{1}{2}\left(\sum\limits_{t=0}^{\infty}\left(G_{w,t}^{\lambda}-V_{w_t}(s_t)\right)^2-\sum\limits_{t=0}^{\infty}\log(\pi_{w_t}\left(A_t\middle|S_t\right))\left(G_{w,t}^{\lambda}-V_{w_t}(s_t)\right)\right)\\
-\psi \sum\limits_{t=0}^{\infty}H_{w_t}(S_t)
\end{multlined}\numberthis\label{meta_objective_entropy}
\end{align*}
Now taking the derivative of equation \ref{meta_objective_entropy}  with respect to $\beta$:\vspace{-3pt}
\begin{align*}
\frac{\partial}{\partial \beta}\mathcal{J}_{\lambda}(w_0..w_{\infty})&=\begin{multlined}[t]\sum\limits_{t=0}^{\infty}\Biggl(\left(\frac{\partial\left(G_{t}^{\lambda}-V_{w_t}(s_t)\right)}{\partial\beta}-\frac{\partial\log(\pi_{w_t}\left(A_t\middle|S_t\right))}{\partial\beta}\right)\left(G_{t}^{\lambda}-V_{w_t}(S_t)\right)\\
-\psi\frac{\partial H_{w_t}(S_t)}{\partial \beta}\Biggr)
\end{multlined}\\
&=-\sum\limits_{t=0}^{\infty}\left(\frac{\partial U_{w_t}(S_t)}{\partial\beta}\left(G_{t}^{\lambda}-V_{w_t}(S_t)\right)+\psi \frac{\partial H_{w_t}(S_t)}{\partial \beta}\right)\\
&=-\sum\limits_{t=0}^{\infty}\left(\frac{\partial U_{w_t}(S_t)}{\partial\beta}\sum\limits_{k=t}^{\infty}(\gamma\lambda)^{k-t}\delta_k+\psi \frac{\partial H_{w_t}(S_t)}{\partial \beta}\right)\\
&=-\sum\limits_{t=0}^{\infty}\begin{multlined}[t]\Biggl(\left<\frac{\partial U_{w_t}(S_t)}{\partial w_t}\middle|\frac{\partial w_t}{\partial \beta}\right>\sum\limits_{k=t}^{\infty}(\gamma\lambda)^{k-t}\delta_k\\
+\psi \left<\frac{\partial H_{w_t}(S_t)}{\partial w_t}\middle|\frac{\partial w_t}{\partial \beta}\right>\Biggr)
\end{multlined}\\
&=\begin{multlined}[t]-\sum\limits_{t=0}^{\infty}\Biggl(\delta_t\sum\limits_{k=0}^{t}(\gamma\lambda)^{t-k}\left<\frac{\partial U_{w_k}(S_k)}{\partial w_k}\middle|\frac{\partial w_k}{\partial \beta}\right>\\
+\psi \left<\frac{\partial H_{w_t}(S_t)}{\partial w_t}\middle|\frac{\partial w_t}{\partial \beta}\right>\Biggr)
\end{multlined}\\
&=-\sum\limits_{t=0}^{\infty}\left( z_{\beta,t}\delta_t+\psi \left<\frac{\partial H_{w_t}(S_t)}{\partial w_t}\middle|\frac{\partial w_t}{\partial \beta}\right>\right)
\end{align*}

We modify $h(t)\dot{=}\frac{\partial w_t}{\partial \beta}$ slightly to account for the entropy regularization. Similar to our handling of the eligibility trace in deriving equation \ref{scalar_h}, we treat all second order derivatives of $H_{w_t}(S_t)$ as zero:\vspace{-3pt}
\begin{align*}
h(t+1)&=\frac{\partial}{\partial \beta}\left(w+\alpha\left(z_t\delta_t+\psi \frac{\partial H_{w_t}(s_t)}{\partial w_t}\right)\right)\\
&\approx h(t)+\alpha\left(z_t\delta_t+\psi \frac{\partial H_{w_t}(s_t)}{\partial w_t}+z_t\frac{\partial \delta_t}{\partial \beta}+\psi\frac{\partial}{\partial\beta}\frac{\partial H_{w_t}(s_t)}{\partial w_t}\right)\\
&=\begin{multlined}[t]h(t)+\alpha\biggl(z_t\delta_t+\psi \frac{\partial H_{w_t}(s_t)}{\partial w_t}+z_t\left<\frac{\partial \delta_t}{\partial w_t}\middle|\frac{\partial w_t}{\partial \beta}\right>\\
+\psi\left<\frac{\partial}{\partial w_t}\frac{\partial H_{w_t}(s_t)}{\partial w_t}\middle|\frac{\partial w_t}{\partial \beta}\right>\biggr)
\end{multlined}\\
&\approx h(t)+\alpha\left(z_t\delta_t+\psi \frac{\partial H_{w_t}(s_t)}{\partial w_t}+z_t\left<\frac{\partial \delta_t}{\partial w_t}\middle|\frac{\partial w_t}{\partial \beta}\right>\right)\\
&=h(t)+\alpha\left(z_t\left(\delta_t+\left<\frac{\partial \delta_t}{\partial w_t}\middle|h(t)\right>\right)+\psi \frac{\partial H_{w_t}(s_t)}{\partial w_t}\right)\\
\end{align*}
All together, Scalar Metatrace for \textit{AC}$(\lambda)$ with entropy regularization is described by:\vspace{-3pt}
\begin{align*}
z_{\beta}&\leftarrow \gamma\lambda z_{\beta}+\left<\frac{\partial U_{w_t}(S_t)}{\partial w_t}\middle|h\right>\\
\beta&\leftarrow \beta+\mu \left(z_{\beta}\delta_t+\psi \left<\frac{\partial H_{w_t}(S_t)}{\partial w_t}\middle|\frac{\partial w_t}{\partial \beta}\right>\right)\\
h&\leftarrow h+e^{\beta}\left(z\left(\delta_t +\left<\gamma\frac{\partial V_{w_t}(S_{t+1})}{\partial w_t}-\frac{\partial V_{w_t}(S_{t})}{\partial w_t}\middle|h\right>\right)+\psi \frac{\partial H_{w_t}(s_t)}{\partial w_t}\right)
\end{align*}
This entropy regularized extension to the basic algorithm is used in lines 7, 10, and 14 of algorithm \ref{alg:NOSMT}. Algorithm \ref{alg:NOSMT} also incorporates a normalization technique, analogous to that used in \cite{autostep}, that we will now discuss.

\subsubsection{Normalization}\label{normalization}
The algorithms discussed so far can be unstable, and sensitive to the parameter $\mu$. Reasons for this and recommended improvements are discussed in \cite{autostep}. We will attempt to map these improvements to our case to improve the stability of our tuning algorithms.

The first issue is that the quantity $\mu z_{\beta}\delta_t$ added to $\beta$ on each time-step is proportional to $\mu$ times a product of $\delta$ values. Depending on the variance of the returns for a particular problem, very different values of $\mu$ may be required to normalize this update. The improvement suggested in \cite{autostep} is straight-forward to map to our case. They divide the $\beta$ update by a running maximum, and we will do the same. This modification to the beta update is done using the factor $v$ on line 9 of algorithm \ref{alg:NOSMT}. $v$ computes a running maximum of the value $\Delta_{\beta}$. $\Delta_{\beta}$ is defined as the value multiplied by $\mu$ to give the update to $\beta$ in the unnormalized algorithm.

The second issue is that updating the step-size by naive gradient descent can rapidly push it into large unstable values (e.g., $\alpha$ larger than one over the squared norm of the feature vector in linear function approximation, which leads to updates which over shoot the target). To correct this they define effective step-size as fraction of distance moved toward the target for a particular update. $\alpha$ is clipped such that effective step-size is at most one (implying the update will not overshoot the target).

With eligibility traces the notion of effective step-size is more subtle. Consider the policy evaluation case and note that our target for a given value function is $G_t^{\lambda}$ and our error is then $\left(G_{t}^{\lambda}-V_{w_t}(s_t)\right)$, the update towards this target is broken into a sum of TD-errors. Nonetheless, for a given fixed step-size our overall update $\Delta V_{w_t}(S_t)$ in the linear case (or its first order approximation in the nonlinear case) to a given value is:
\begin{equation}\label{scalarupdate}
\Delta V_{w_t}(S_t)=\alpha\left|\frac{\partial V_{w_t}(S_{t})}{\partial w_t}\right|^2\left(G_{t}^{\lambda}-V_{w_t}(s_t)\right)
\end{equation}
Dividing by the error, our fractional update, or ``effective step-size'', is then:
\begin{equation*}
\alpha\left|\frac{\partial V_{w_t}(S_{t})}{\partial w_t}\right|^2
\end{equation*}
However, due to our use of eligibility traces, we will not be able to update each state's value function with an independent step-size but must update towards the partial target for multiple states at each timestep with a shared step-size. A reasonable way to proceed then would be to simply choose our shared (scalar or vector) $\alpha$ to ensure that the maximum effective-step size for any state contributing to the current trace is still less than one to avoid overshooting. We maintain a running maximum of effective step-sizes over states in the trace, similar to the running maximum of $\beta$ updates used in \cite{autostep}, and multiplicatively decaying our $\alpha$s on each time-step by the amount this exceeds one. This procedure is shown on lines 11, 12 and 13 of algorithm \ref{alg:NOSMT}.

Recall that the effective step-size bounding procedure we just derived was for the case of policy evaluation. For the AC($\lambda$), this is less clear as it is not obvious what target we should avoid overshooting with the policy parameters. We simply replace $\frac{\partial V_{w_t}(s_t)}{\partial w_t}$ with
$\left(\frac{\partial V_{w_t}(s_t)}{\partial w_t}+\frac{1}{2}\frac{\partial \log(\pi_{w_t}\left(A_t\middle|S_t\right))}{\partial w_t}\right)$ in the computation of $u$. This is a conservative heuristic which normalizes $\alpha$ as if the combined policy improvement, policy evaluation objective were a pure policy evaluation objective. 

We consider this reasonable since we know of no straightforward way to place an upper bound on the useful step-size for policy improvement but it nonetheless makes sense to make sure it does not grow arbitrarily large. The constraint on the $\alpha$s for policy evaluation will always be tighter than constraining based on the value derivatives alone. Other options for normalizing the step-size may be worth considering as well.
We will depart from \cite{autostep} somewhat by choosing $\mu$ itself as the tracking parameter for $v$ rather than $\frac{1}{\tau}\alpha_i\frac{\partial V_{w_t}(S_{t+1})}{\partial w_t}^2$, where $\tau$ is a hyperparameter. This is because there is no obvious reason to use $\alpha_i\frac{\partial V_{w_t}(S_{t+1})}{\partial w_t}^2$ specifically here, and it is not clear whether the appropriate analogy for the RL case is to use $\alpha_i\frac{\partial V_{w_t}(S_{t+1})}{\partial w_t}^2$, $\alpha_i z_t^2$, or something else. We use $\mu$ for simplicity, thus enforcing the tracking rate of the maximum used to normalize the step-size updates be proportional to the the magnitude of step-size updates. For the tracking parameter of $u$ we choose $(1-\gamma\lambda)$, roughly taking the maximum over all the states that are currently contributing to the trace, which is exactly what we want $u$ to track.

\begin{algorithm}[t]
\begin{algorithmic}[1]
\State $h\gets 0$, $\beta\gets \beta_0$, $v\gets 0$
\For{each episode}
\State $z_{\beta}\gets 0$, $u\gets 0$
\While{episode not complete}
\State \textbf{receive } $V_{w_t}(S_t), \pi_{w_t}(S_t|A_t), \delta_t, z$
\State $U_{w_t} \gets V_{w_t}+\frac{1}{2}\log(\pi_{w_t}\left(A_t\middle|S_t\right))$
\State $z_{\beta}\gets \gamma\lambda z_{\beta}+\left<\frac{\partial U_{w_t}}{\partial w_t}\middle| h\right>$
\State $\Delta_{\beta}\gets z_{\beta}\delta_t+\psi\left<\frac{\partial H_{w_t}(S_t)}{\partial w_t}\middle| h\right>$
\State $v\gets\max(|\Delta_{\beta}|,v+\mu(|\Delta_{\beta}|-v)$
\State $\beta\gets \beta+\mu \frac{\Delta_{\beta}}{\left(v\text{ if } v>0\text{ else }1\right)}$
\State $u\gets \max(e^{\beta}\left|\frac{\partial U_{w_t}}{\partial w_t}\right|^2,u+(1-\gamma\lambda)(e^{\beta}\left|\frac{\partial U_{w_t}}{\partial w_t}\right|^2-u))$
\State $M\gets \max\left(u,1\right)$
\State $\beta\gets \beta-\log(M)$
\State $h\gets h+e^{\beta} (z(\delta_t+\left<\frac{\partial\delta_t}{\partial w_t}\middle|h\right>)+\psi\frac{\partial H_{w_t}(S_t)}{\partial w_t})$
\State \textbf{output } $\alpha=e^{\beta}$
\EndWhile
\EndFor
\end{algorithmic}
\caption{Normalized Scalar Metatrace for Actor-Critic}
\label{alg:NOSMT}
\end{algorithm}
\subsection{Vector Metatrace for \textit{AC}$(\lambda)$}
Now define $\alpha_i=e^{\beta_i}$ to be the $i^{th}$ element of a vector of step-sizes, one for each weight such that the update for each weight element in \textit{AC}$(\lambda)$ will use the associated $\alpha_i$. Having a separate $\alpha_i$ for each weight enables the algorithm to individually adjust how quickly it tracks each feature. This becomes particularly important when the state representation is non-stationary, as is the case when NN based models are used. In this case, features may be changing at different rates, and some may be much more useful than others, we would like our algorithm to be able to assign high step-size to fast changing useful features while annealing the step-size of features that are either mostly stationary or not useful to avoid tracking noise \cite{TIDBD,idbd}.

Take each $w_t$ to be a function of $\beta_i$ for all $i$ and following \cite{idbd} use the approximation $\frac{\partial w_{i,t}}{\partial\beta_j}=0$ for all $i\neq j$, differentiate equation \ref{control_objective}, again with the same semi-gradient treatment used in \textit{AC}$(\lambda)$ to yield:\vspace{-3pt}
\begin{align*}
\frac{\partial}{\partial \beta_i}\mathcal{J}^{\beta}_{\lambda}(w_0..w_{\infty})&=-\sum\limits_{t=0}^{\infty}\frac{\partial U_{w_t}(S_t)}{\partial\beta_i}\left(G_{t}^{\lambda}-V_{w_t}(S_t)\right)\\
&\approx -\sum\limits_{t=0}^{\infty}\frac{\partial U_{w_t}(S_t)}{\partial w_{i,t}}\frac{\partial w_{i,t}}{\partial \beta_i}\sum\limits_{k=t}^{\infty}(\gamma\lambda)^{k-t}\delta_k\\
&=-\sum\limits_{t=0}^{\infty}\delta_t\sum\limits_{k=0}^{t}(\gamma\lambda)^{t-k}\frac{\partial U_{w_k}(S_k)}{\partial w_{i,k}}\frac{\partial w_{i,k}}{\partial \beta_i}
\end{align*}
Once again define an eligibility trace, now a vector, with elements:\vspace{-3pt}
\begin{equation*}
z_{\beta,i,t}\dot{=}\sum\limits_{k=0}^{t}(\gamma\lambda)^{t-k}\frac{\partial U_{w_k}(S_k)}{\partial w_{i,k}}\frac{\partial w_{i,k}}{\partial \beta_i}
\end{equation*}
such that:\vspace{-3pt}
\begin{equation*}
\frac{\partial}{\partial \beta_i}\mathcal{J}_{\lambda}(w_0..w_{\infty})\approx-\sum\limits_{t=0}^{\infty}\delta_tz_{\beta,i,t}
\end{equation*}
To compute $h_i(t)\dot{=}\frac{\partial w_{i,t}}{\partial \beta_i}$ we use the obvious generalization of the scalar case, again we use the approximation $\frac{\partial w_{i,t}}{\partial\beta_j}=0$ for all $i\neq j$:\vspace{-3pt}
\begin{align*}
h_i(t+1)&\approx h_i(t)+\alpha_i z_{i,t}\left(\delta_t +\left<\frac{\partial \delta_t}{\partial w_t}\middle|\frac{\partial w_{t}}{\partial \beta_i}\right>\right)\\
&\approx h_i(t)+\alpha_i z_{i,t}\left(\delta_t +\frac{\partial \delta_t}{\partial w_{i,t}}\frac{\partial w_{i,t}}{\partial \beta_i}\right)\\
&=h_i(t)+\alpha_iz_{i,t}\left(\delta_t+\left(\gamma\frac{\partial V_{w_t}(S_{t+1})}{\partial {w_{i,t}}}-\frac{\partial V_{w_t}(S_{t})}{\partial {w_{i,t}}}\right)h_i(t)\right)
\end{align*}
The first line approximates $\frac{\partial z_{i,t}}{w_t}=0$ as in the scalar case. The second line approximates $\frac{\partial w_{i,t}}{\partial\beta_j}=0$ as discussed above. All together, Vector Metatrace for \textit{AC}$(\lambda)$  is described by:\vspace{-3pt}
\begin{align*}
z_{\beta}&\leftarrow \gamma\lambda z_{\beta}+\frac{\partial U_{w_t}(S_t)}{\partial w_t}\odot h\\
\beta&\leftarrow \beta+\mu z_{\beta}\delta_t\\
h&\leftarrow h+e^{\beta}\odot z\odot\left(\delta_t+\left(\gamma\frac{\partial V_{w_t}(S_{t+1})}{\partial w_t}-\frac{\partial V_{w_t}(S_{t})}{\partial w_t}\right)\odot h\right)
\end{align*}
where $\odot$ denotes element-wise multiplication. 

As in the scalar case we augment this basic algorithm with entropy regularization and normalization. The extension of entropy regularization to the vector case is a straightforward modification of the derivation presented in Section \ref{entropy_reg}. To extend the normalization technique note that $v$ is now replaced by a vector denoted $\overrightarrow{v}$ on line 9 of algorithm \ref{alg:NOVMT}. $\overrightarrow{v}$ maintains a separate running maximum for the updates to each element of $\overrightarrow{\beta}$. To extend the notion of effective step-size to the vector case, follow the same logic used to derive equation \ref{scalarupdate} to yield an update $\Delta V_{w_t}(S_t)$ of the form:
\begin{equation}
\Delta V_{w_t}(S_t)=\left<\alpha\middle|\frac{\partial V_{w_t}(S_{t})}{\partial w_t}^2\right>\left(G_{t}^{\lambda}-V_{w_t}(s_t)\right)
\end{equation}
where $\alpha$ is now vector valued. Dividing by the error our fractional update, or ``effective step-size'', is then:
\begin{equation}
\left<\alpha\middle|\frac{\partial V_{w_t}(S_{t})}{\partial w_t}^2\right>
\end{equation}
As in the scalar case, we maintain a running maximum of effective step-sizes over states in the trace and multiplicatively decay $\alpha$ on each time step by the amount this exceeds one. This procedure is shown on lines 11, 12 and 13 of algorithm \ref{alg:NOVMT}. The full algorithm for the vector case augmented with entropy regularization and normalization is presented in algorithm \ref{alg:NOVMT}.
\begin{algorithm}[t]
\begin{algorithmic}[1]
\State $h\gets 0$, $\overrightarrow{\beta}\gets \beta_0$, $\overrightarrow{v}\gets 0$
\For{each episode}
\State $\overrightarrow{z_{\beta}}\gets 0$, $u\gets 0$
\While{episode not complete}
\State \textbf{receive } $V_{w_t}(S_t), \pi_{w_t}(A_t|S_t), \delta_t, z$
\State $U_{w_t} \gets V_{w_t}+\frac{1}{2}\log(\pi_{w_t}\left(A_t\middle|S_t\right))$
\State $\overrightarrow{z_{\beta}}\gets \gamma\lambda \overrightarrow{z_{\beta}}+\frac{\partial U_{w_t}}{\partial w_t}\odot \overrightarrow{h}$
\State $\Delta_{\overrightarrow{\beta}}\gets \overrightarrow{z_{\beta}}\delta_t+\psi\frac{\partial H_{w_t}(S_t)}{\partial w_t}\odot \overrightarrow{h}$
\State $\overrightarrow{v}\gets\max(|\Delta_{\overrightarrow{\beta}}|,\overrightarrow{v}+\mu(|\Delta_{\overrightarrow{\beta}}|-\overrightarrow{v})$
\State $\overrightarrow{\beta}\gets \overrightarrow{\beta}+\mu \frac{\Delta_{\overrightarrow{\beta}}}{\left(\overrightarrow{v}\text{ where } \overrightarrow{v}>0\text{ elsewhere }1\right)}$
\State $u\gets \max\left(\left<e^{\overrightarrow{\beta}}\middle|\frac{\partial U_{w_t}}{\partial w_t}^2\right>,u+(1-\gamma\lambda)\left(\left<e^{\overrightarrow{\beta}}\middle|\frac{\partial U_{w_t}}{\partial w_t}^2\right>-u\right)\right)$
\State $M\gets \max\left(u,1\right)$
\State $\overrightarrow{\beta}\gets \overrightarrow{\beta}-\log(M)$
\State $\overrightarrow{h}\gets \overrightarrow{h}+e^{\overrightarrow{\beta}}\odot\left( \overrightarrow{z}\odot\left(\delta_t+\frac{\partial\delta_t}{\partial w_t}\odot \overrightarrow{h}\right)+\psi\frac{\partial H_{w_t}(S_t)}{\partial w_t}\right)$
\State \textbf{output } $\overrightarrow{\alpha}=e^{\overrightarrow{\beta}}$
\EndWhile
\EndFor
\end{algorithmic}
\caption{Normalized Vector Metatrace for Actor-Critic}
\label{alg:NOVMT}
\end{algorithm}
\subsection{Mixed Metatrace}
We also explore a mixed algorithm where a vector correction to the step-size $\overrightarrow{\beta}$ is learned for each weight and added to a global value $\hat{\beta}$ which is learned collectively for all weights. The derivation of this case is a straightforward extension of the scalar and vector cases. The full algorithm, that also includes entropy regularization and normalization, is detailed in Algorithm \ref{alg:NOMMT}.
\begin{algorithm}
\begin{algorithmic}[1]
\State $\overrightarrow{h}\gets 0$, $\hat{h}\gets 0$, $\hat{\beta}\gets \beta_0$, $\overrightarrow{\beta} \gets 0$, $\hat{v}\gets 0$, $\overrightarrow{v}\gets 0$
\For{each episode}
\State $\overrightarrow{z_{\beta}}\gets 0$, $\hat{z_{\beta}}\gets 0$, $u\gets 0$
\While{episode not complete}
\State \textbf{receive } $V_{w_t}(S_t), \pi_{w_t}(S_t|A_t), \delta_t, z$
\State $U_{w_t} \gets V_{w_t}+\frac{1}{2}\log(\pi_{w_t}\left(A_t\middle|s_t\right))$
\State $\overrightarrow{z_{\beta}}\gets \gamma\lambda \overrightarrow{z_{\beta}}+\frac{\partial U_{w_t}}{\partial w_t}\odot \overrightarrow{h}$
\State $\hat{z_{\beta}}\gets \gamma\lambda \hat{z_{\beta}}+\left<\frac{\partial U_{w_t}}{\partial w_t}\middle| \hat{h}\right>$
\State $\Delta_{\overrightarrow{\beta}}\gets \overrightarrow{z_{\beta}}\delta_t+\psi\frac{\partial H_{w_t}(S_t)}{\partial w_t}\odot \overrightarrow{h}$
\State $\Delta_{\hat{\beta}}\gets \hat{z_{\beta}}\delta_t+\psi\left<\frac{\partial H_{w_t}(S_t)}{\partial w_t}\middle| \hat{h}\right>$
\State $\overrightarrow{v}\gets\max(|\Delta_{\overrightarrow{\beta}}|,\overrightarrow{v}+\mu(|\Delta_{\overrightarrow{\beta}}|-\overrightarrow{v})$
\State $\hat{v}\gets\max(|\Delta_{\hat{\beta}}|,\hat{v}+\mu(|\Delta_{\hat{\beta}}|-\hat{v})$
\State $\overrightarrow{\beta}\gets \overrightarrow{\beta}+\mu \frac{\Delta_{\overrightarrow{\beta}}}{\left(\overrightarrow{v}\text{ where } v>0\text{ elsewhere }1\right)}$
\State $\hat{\beta}\gets \hat{\beta}+\mu \frac{\Delta_{\hat{\beta}}}{\left(\hat{v}\text{ if } \hat{v}>0\text{ else }1\right)}$
\State $u\gets \max\left(\left<\exp(\hat{\beta}+\overrightarrow{\beta})\middle|\frac{\partial U_{w_t}}{\partial w_t}^2\right>,u+(1-\gamma\lambda)\left(\left<\exp(\hat{\beta}+\overrightarrow{\beta})\middle|\frac{\partial U_{w_t}}{\partial w_t}^2\right>-u\right)\right)$
\State $M\gets \max\left(u,1\right)$
\State $\hat{\beta}\gets \hat{\beta}-\log(M)$
\State $\overrightarrow{h}\gets \overrightarrow{h}+\exp(\hat{\beta}+\overrightarrow{\beta})\odot\left( z\odot\left(\delta_t+\frac{\partial\delta_t}{\partial w_t}\odot \overrightarrow{h}\right)+\psi\frac{\partial H_{w_t}(S_t)}{\partial w_t}\right)$
\State $\hat{h}\gets \hat{h}+\exp(\hat{\beta}+\overrightarrow{\beta})\odot\left(z\left(\delta_t+\left<\frac{\partial\delta_t}{\partial w_t}\middle|\hat{h}\right>\right) +\psi\frac{\partial H_{w_t}(S_t)}{\partial w_t}\right)$
\State \textbf{output } $\overrightarrow{\alpha}=\exp(\hat{\beta}+\overrightarrow{\beta})$
\EndWhile
\EndFor
\end{algorithmic}
\caption{Normalized Mixed Metatrace for Actor-Critic}
\label{alg:NOMMT}
\end{algorithm}

\section{Experiments and Results}
\subsection{Mountain Car}
We begin by testing Scalar Metatrace on the classic mountain car domain using the implementation available in OpenAI gym \cite{gym}. A reward of $-1$ is given for each time-step until the goal is reached or a maximum of $200$ steps are taken, after which the episode terminates. We use tile-coding for state representation: tile-coding with 16 10x10 tilings creates a feature vector of size $1600$. The learning algorithm is \textit{AC}$(\lambda)$ with $\gamma$ fixed to $0.99$ and $\lambda$ fixed to $0.8$ in all experiments. For mountain car we do not use any entropy regularization, and all weights were initialized to 0.

\begin{figure}[!t]
\begin{center}
\begin{subfigure}[t]{0.44\textwidth}
\vskip 0pt
\includegraphics[width=\textwidth]{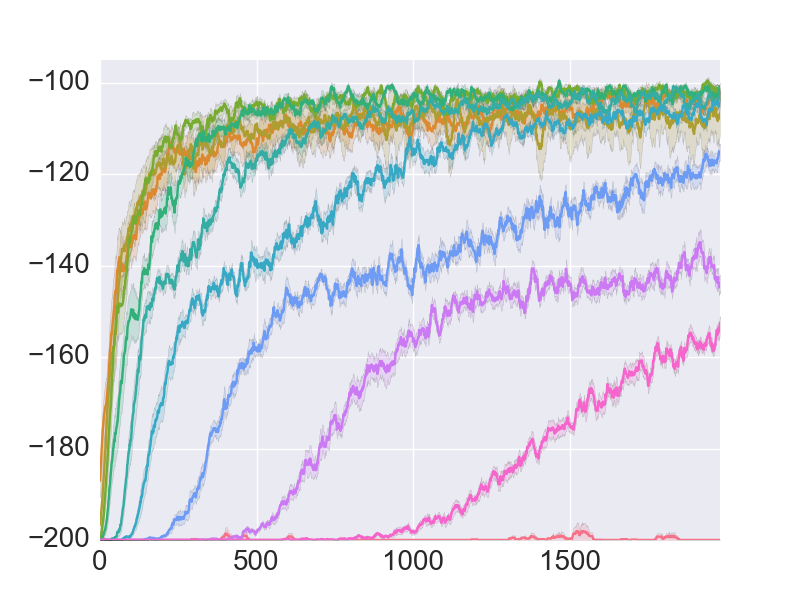}
\end{subfigure}
\begin{subfigure}[t]{0.10\textwidth}
\vskip 10pt
\includegraphics[width=\textwidth]{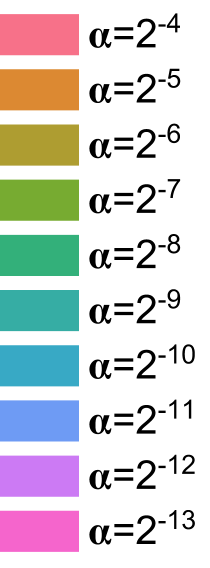}
\end{subfigure}
\end{center}\vspace{-15pt}
\caption{Return vs. training episodes on mountain car for a variety of $\alpha$ values with no step-size tuning. Each curve shows the average of 10 repeats and is smoothed by taking the average of the last 20 episodes.}
\label{MC_baseline}
\end{figure}

\begin{figure}[!t]
\begin{subfigure}[t]{0.44\textwidth}
\vskip 0pt
\includegraphics[width=\textwidth]{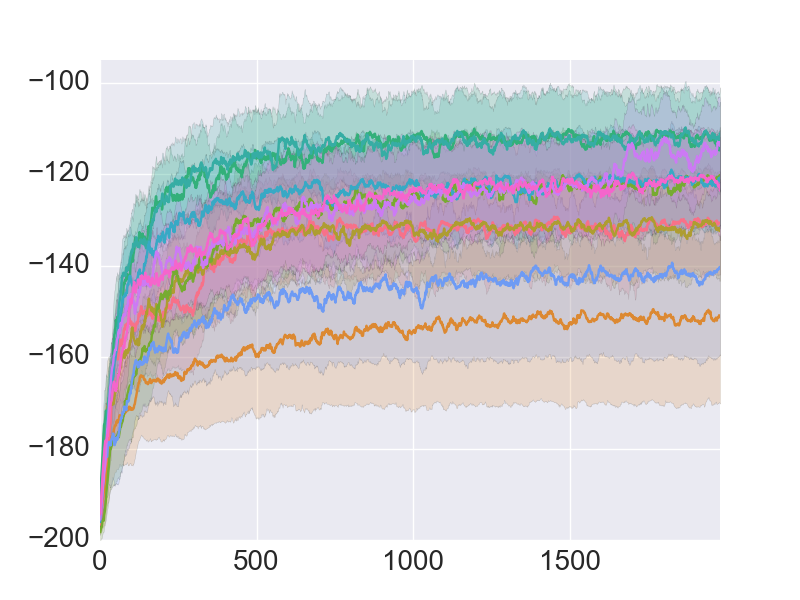}
\caption{$\mu=2^{-6}$}
\end{subfigure}
\begin{subfigure}[t]{0.10\textwidth}
\vskip 10pt
\includegraphics[width=\textwidth]{alpha_legend.png}
\end{subfigure}
\begin{subfigure}[t]{0.44\textwidth}
\vskip 0pt
\includegraphics[width=\textwidth]{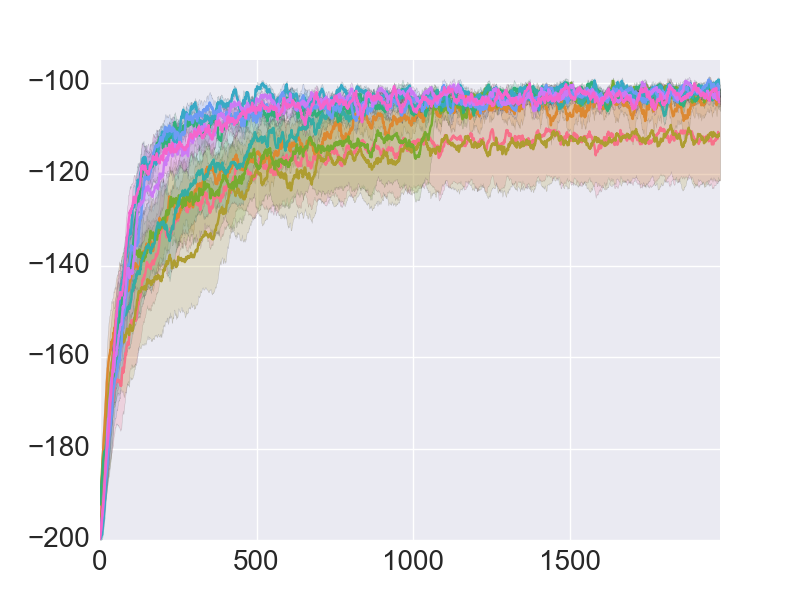}
\caption{$\mu=2^{-7}$}
\end{subfigure}
\linebreak
\begin{subfigure}[t]{0.44\textwidth}
\includegraphics[width=\textwidth]{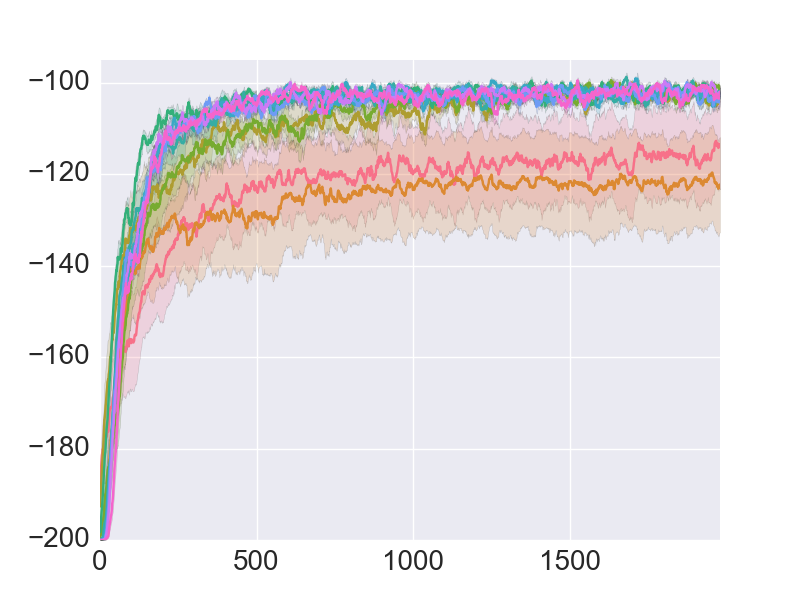}
\caption{$\mu=2^{-8}$}
\end{subfigure}
\hfill
\begin{subfigure}[t]{0.44\textwidth}
\includegraphics[width=\textwidth]{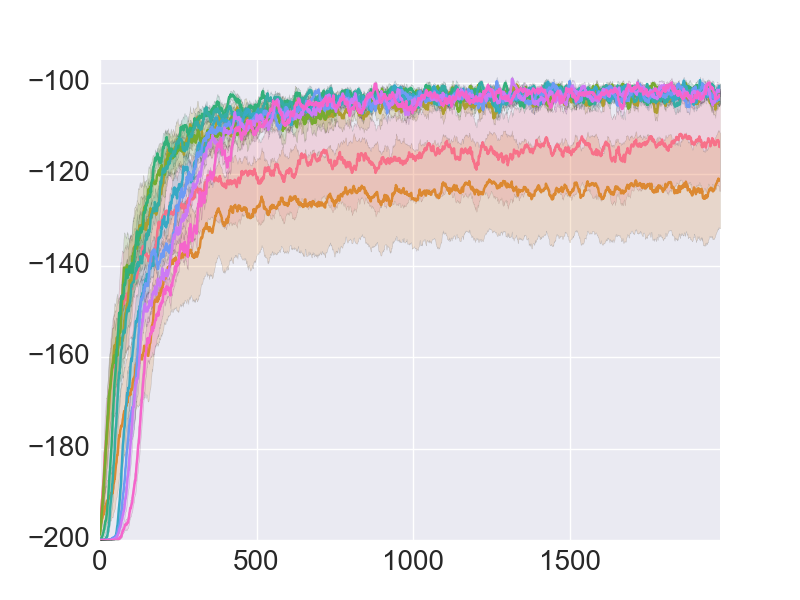}
\caption{$\mu=2^{-9}$}
\end{subfigure}
\linebreak
\begin{subfigure}[t]{0.44\textwidth}
\includegraphics[width=\textwidth]{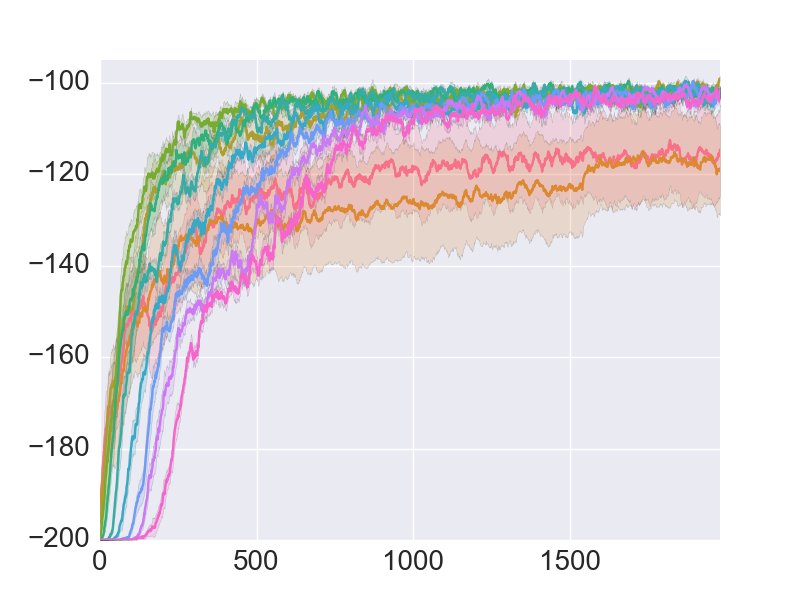}
\caption{$\mu=2^{-10}$}
\end{subfigure}
\hfill
\begin{subfigure}[t]{0.44\textwidth}
\includegraphics[width=\textwidth]{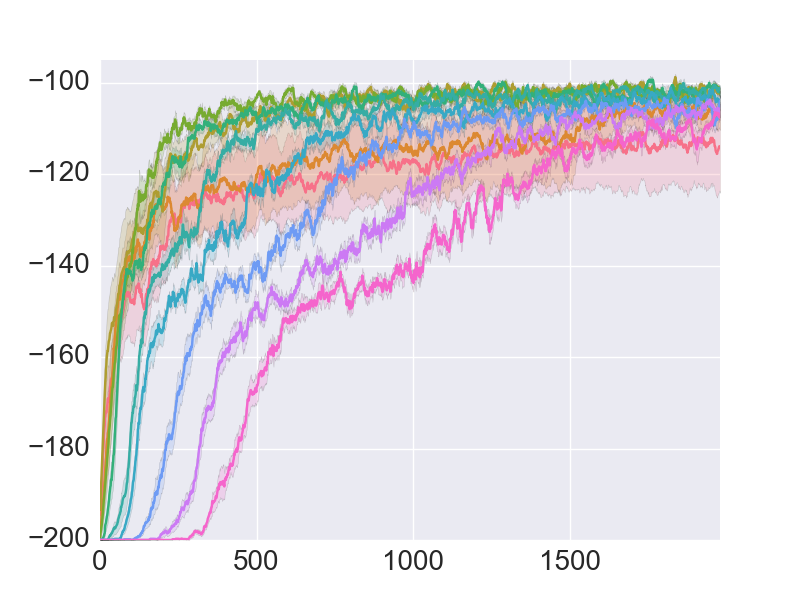}
\caption{$\mu=2^{-11}$}
\end{subfigure}
\caption{Return vs. training episodes on mountain car for a variety of initial $\alpha$ values with Normalized Scalar Metatrace tuning with a variety of $\mu$ values. Each curve shows the average of 10 repeats and is smoothed by taking the average of the last 20 episodes. For a broad range of $\mu$ values, scalar Metatrace is able to accelerate learning, particularly for suboptimal values of the initial $\alpha$.}
\label{MC_scalar_normalized}
\end{figure}

Figure \ref{MC_baseline} shows the results on this problem for a range of $\alpha$ values without step-size tuning. $\alpha$ values were chosen as powers of $2$ which range from excessively low to too high to allow learning. Figure \ref{MC_scalar_normalized} shows the results of Normalized Scalar Metatrace for a range of $\mu$ values. For a fairly broad range of $\mu$ values, learning curves of different $\alpha$ values are much more similar compared to without tuning.
%\FloatBarrier

\begin{figure}[!t]
\begin{subfigure}[t]{0.44\textwidth}
\vskip 0pt
\includegraphics[width=\textwidth]{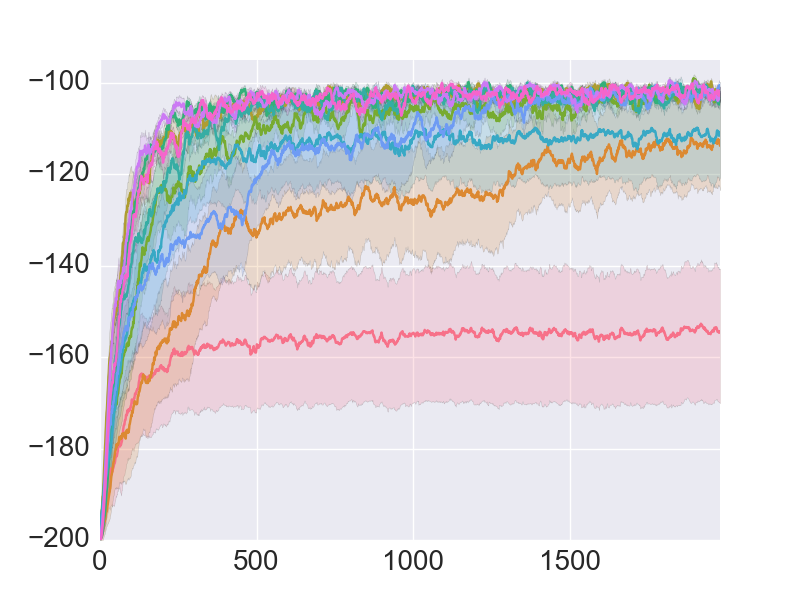}
\caption{$\mu=2^{-14}$}
\end{subfigure}
\begin{subfigure}[t]{0.12\textwidth}
\vskip 10pt
\includegraphics[width=\textwidth]{alpha_legend.png}
\end{subfigure}
\begin{subfigure}[t]{0.44\textwidth}
\vskip 0pt
\includegraphics[width=\textwidth]{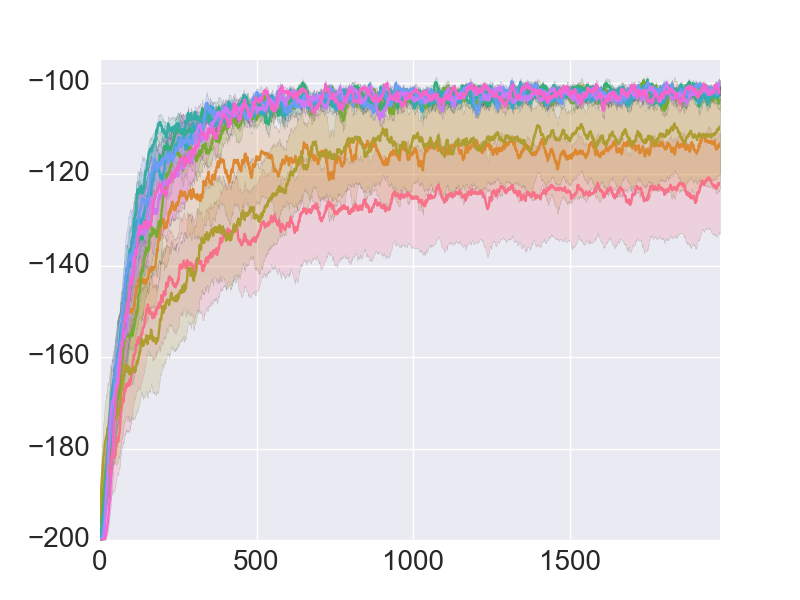}
\caption{$\mu=2^{-15}$}
\end{subfigure}
\linebreak
\begin{subfigure}[t]{0.44\textwidth}
\includegraphics[width=\textwidth]{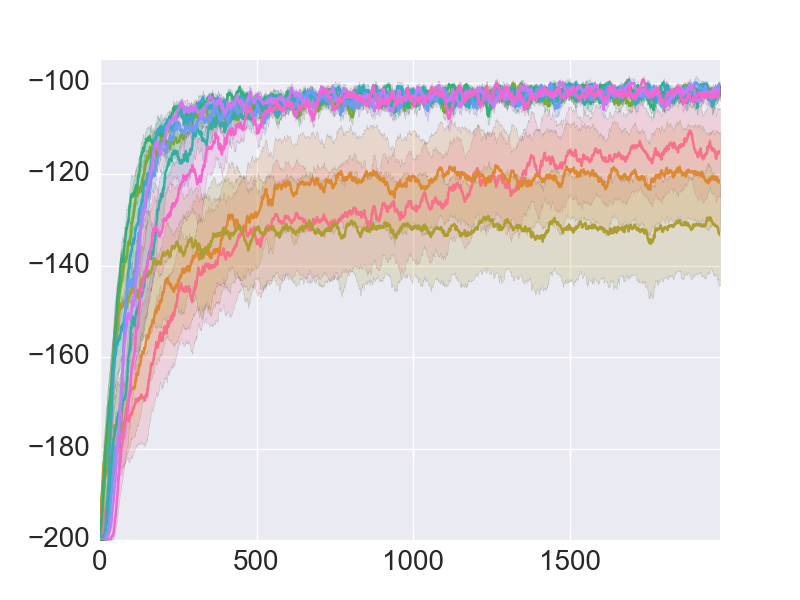}
\caption{$\mu=2^{-16}$}
\end{subfigure}
\hfill
\begin{subfigure}[t]{0.44\textwidth}
\includegraphics[width=\textwidth]{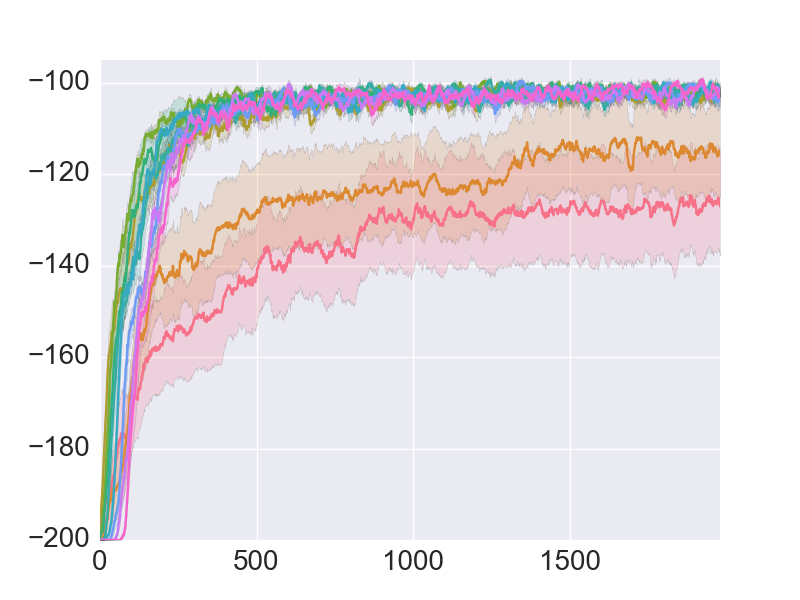}
\caption{$\mu=2^{-17}$}
\end{subfigure}
\linebreak
\begin{subfigure}[t]{0.44\textwidth}
\includegraphics[width=\textwidth]{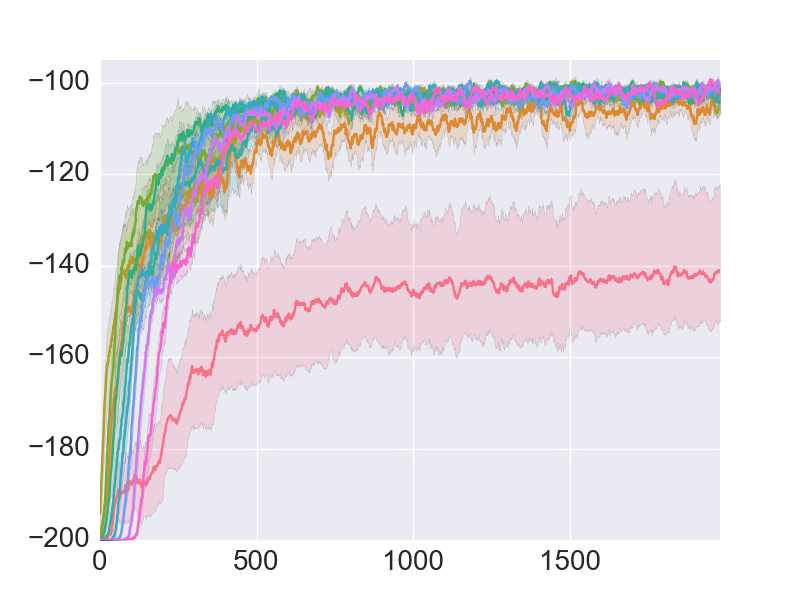}
\caption{$\mu=2^{-18}$}
\end{subfigure}
\hfill
\begin{subfigure}[t]{0.44\textwidth}
\includegraphics[width=\textwidth]{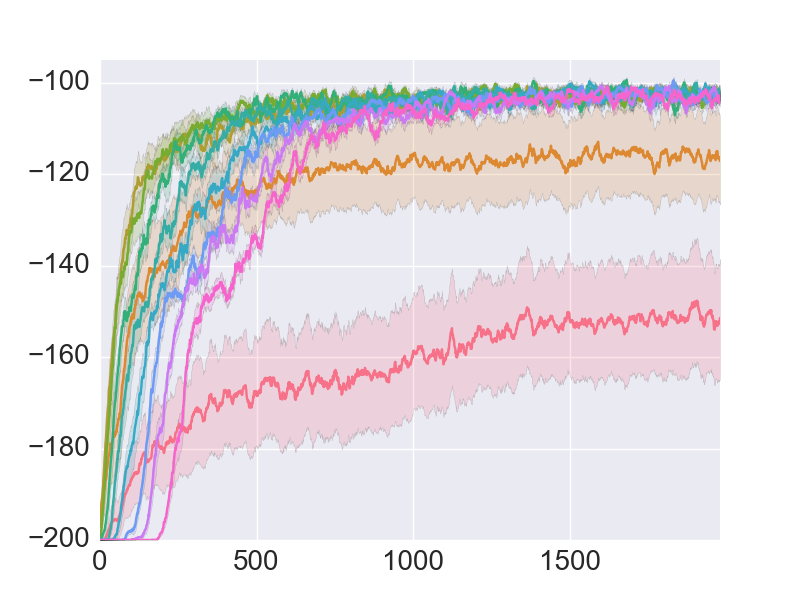}
\caption{$\mu=2^{-19}$}
\end{subfigure}
\caption{Return on mountain-car for a variety of initial $\alpha$ values with unnormalized scalar Metatrace with a variety of $\mu$ values. Each curve shows the average of 10 repeats and is smoothed by taking the average of the last 20 episodes.}
\label{MC_scalar_unnormalized}
\end{figure}

Figure \ref{MC_scalar_unnormalized} shows the performance of Unnormalized Scalar Metatrace over a range of $\mu$ and initial $\alpha$ values. As in Figure 2 for Normalized Scalar Metatrace, for a fairly broad range of $\mu$ values, learning curves over different initial $\alpha$ values are much more similar compared to without tuning. Two qualitative differences of the normalized algorithm over the unnormalized version are: (1) the useful $\mu$ values are much larger, which for reasons outlined in \cite{idbd} likely reflects less problem dependence, and (2) the highest initial alpha value tested, $2^{-5}$, which does not improve at all without tuning, becomes generally more stable. In both the normalized and unnormalized case, higher $\mu$ values than those shown cause increasing instability.
% \FloatBarrier

\begin{figure}[!t]
\begin{subfigure}[t]{0.44\textwidth}
\vskip 0pt
\includegraphics[width=\textwidth]{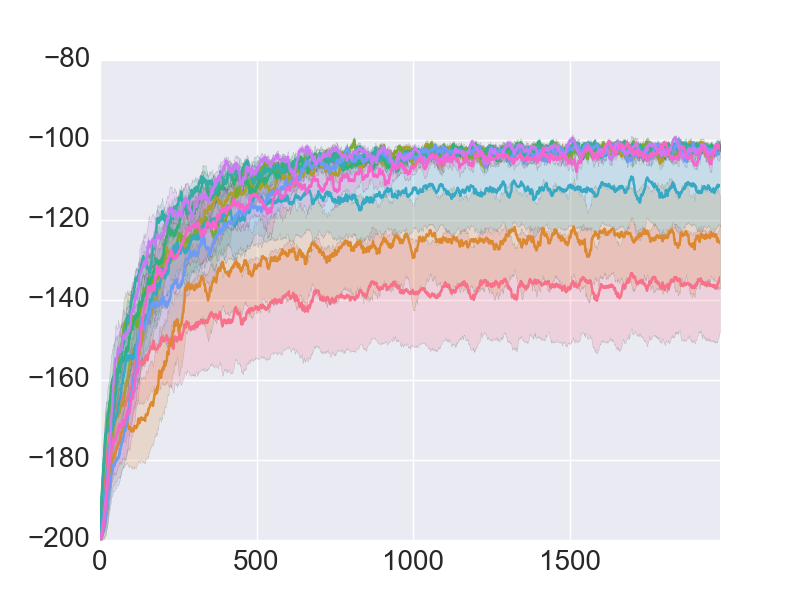}
\caption{$\mu=2^{-14}$}
\end{subfigure}
\begin{subfigure}[t]{0.12\textwidth}
\vskip 10pt
\includegraphics[width=\textwidth]{alpha_legend.png}
\end{subfigure}
\begin{subfigure}[t]{0.44\textwidth}
\vskip 0pt
\includegraphics[width=\textwidth]{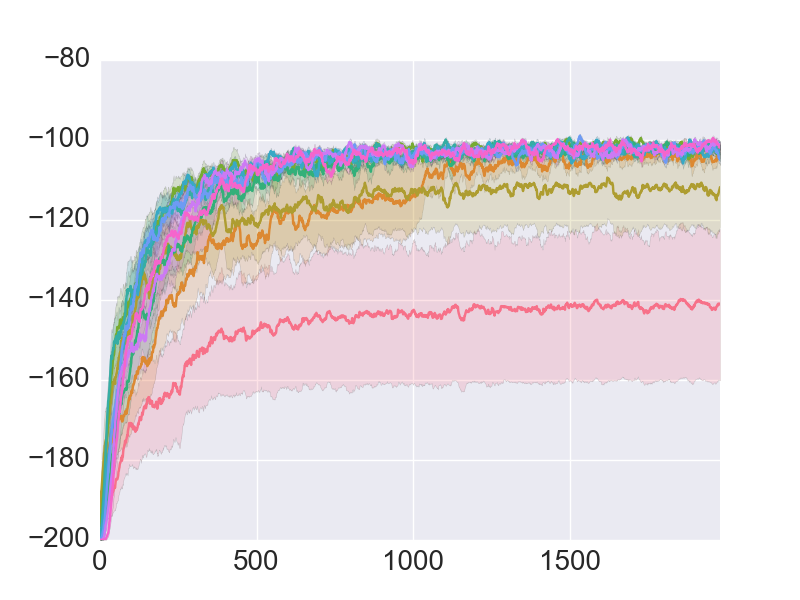}
\caption{$\mu=2^{-15}$}
\end{subfigure}
\linebreak
\begin{subfigure}[t]{0.44\textwidth}
\includegraphics[width=\textwidth]{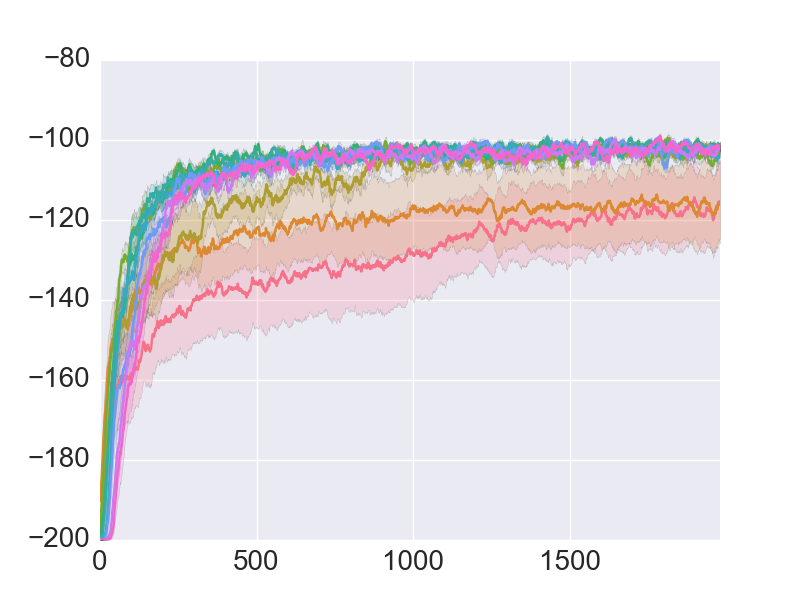}
\caption{$\mu=2^{-16}$}
\end{subfigure}
\hfill
\begin{subfigure}[t]{0.44\textwidth}
\includegraphics[width=\textwidth]{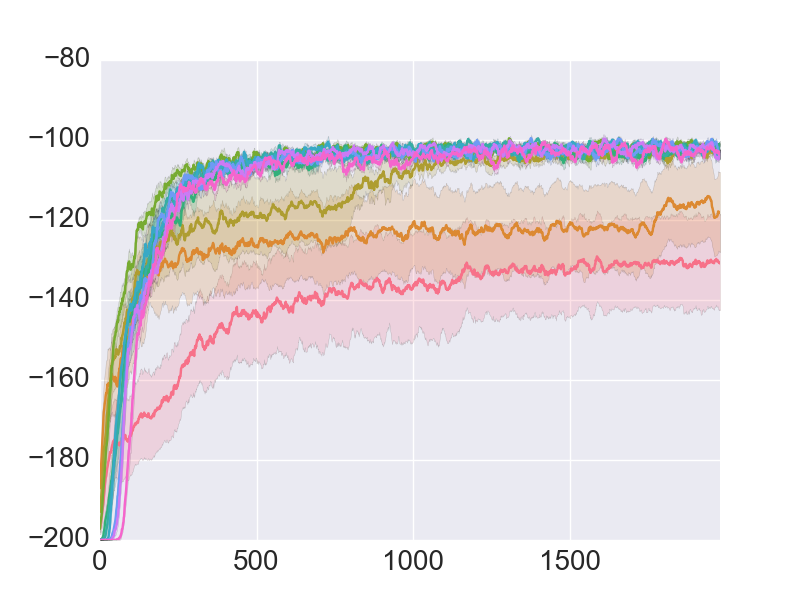}
\caption{$\mu=2^{-17}$}
\end{subfigure}
\linebreak
\begin{subfigure}[t]{0.44\textwidth}
\includegraphics[width=\textwidth]{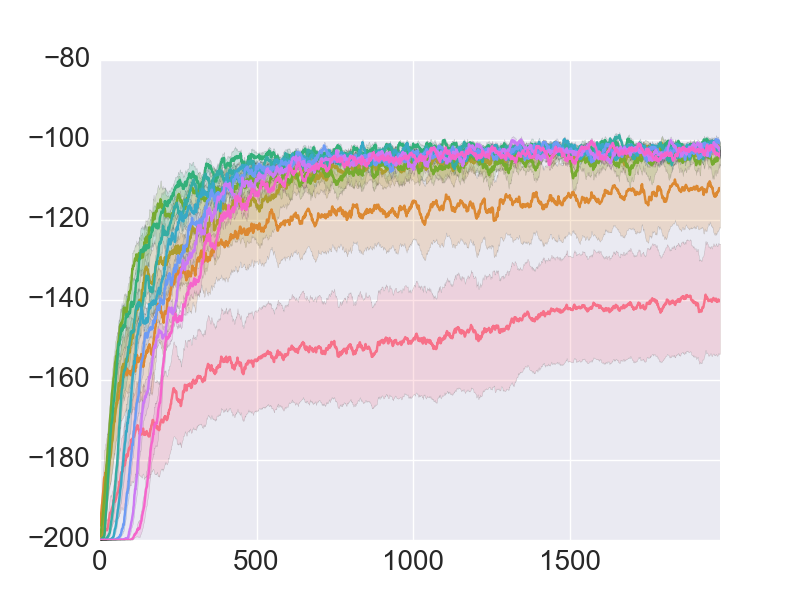}
\caption{$\mu=2^{-18}$}
\end{subfigure}
\hfill
\begin{subfigure}[t]{0.44\textwidth}
\includegraphics[width=\textwidth]{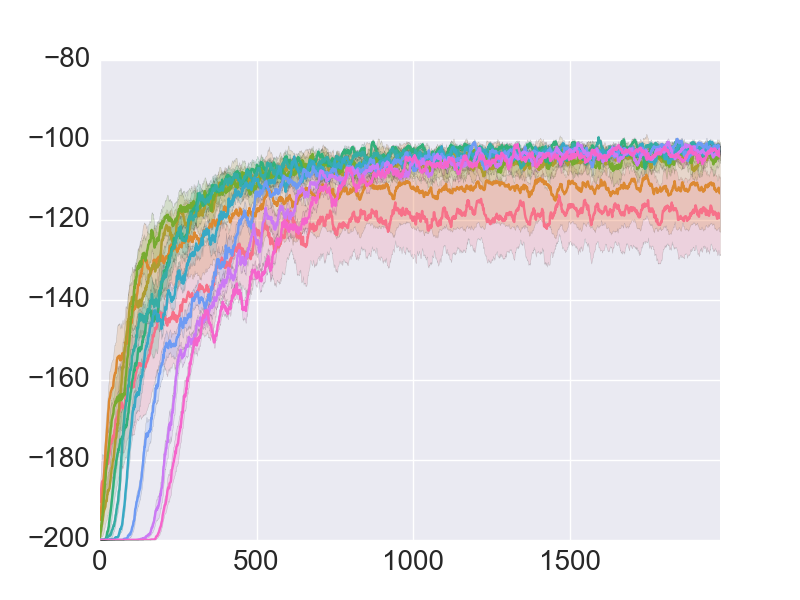}
\caption{$\mu=2^{-19}$}
\end{subfigure}
\caption{Return on mountain-car for a variety of initial $\alpha$ values with SID with a variety of $\mu$ values. Each curve shows the average of 10 repeats and is smoothed by taking the average of the last 20 episodes. On this problem the behavior of SID is very similar to that of Scalar Metatrace.}
\label{SID}
\end{figure}

\begin{figure}[!t]
\begin{subfigure}[t]{0.44\textwidth}
\vskip 0pt
\includegraphics[width=\textwidth]{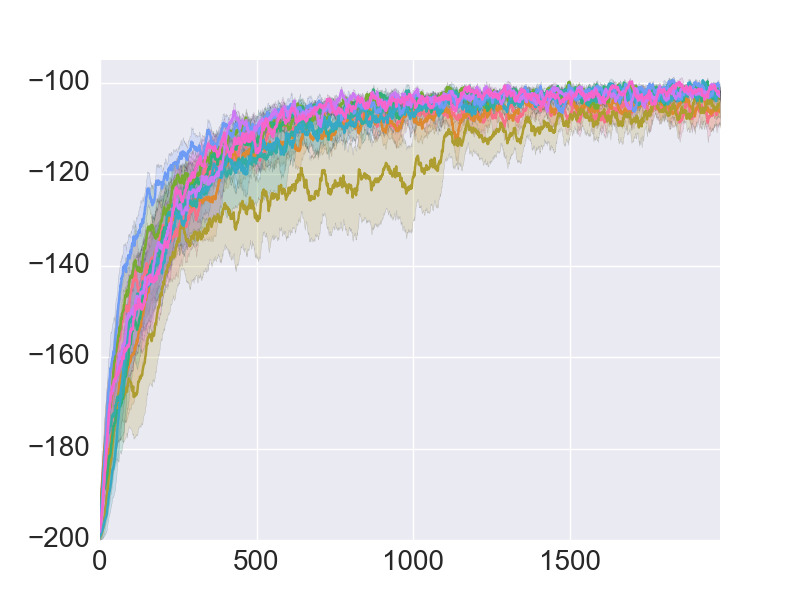}
\caption{$\mu=2^{-10}$}
\end{subfigure}
\begin{subfigure}[t]{0.12\textwidth}
\vskip 10pt
\includegraphics[width=\textwidth]{alpha_legend.png}
\end{subfigure}
\begin{subfigure}[t]{0.44\textwidth}
\vskip 0pt
\includegraphics[width=\textwidth]{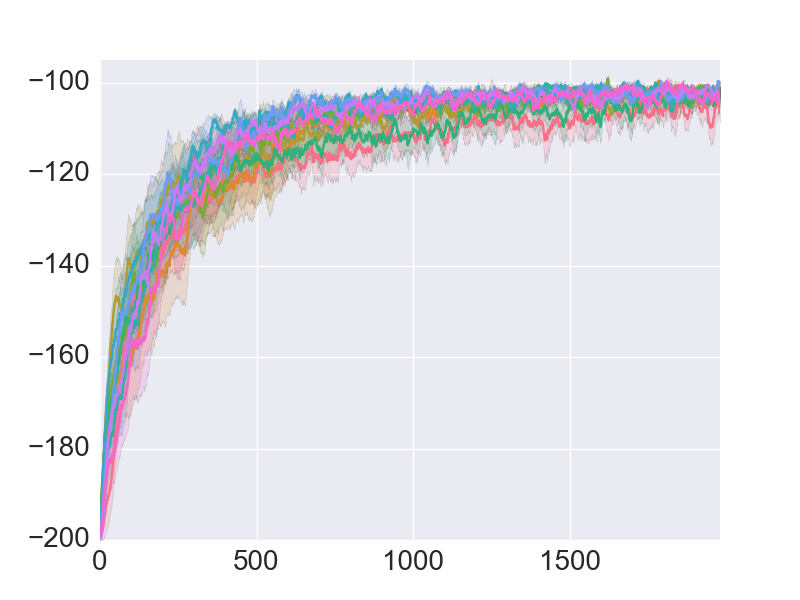}
\caption{$\mu=2^{-11}$}
\end{subfigure}
\linebreak
\begin{subfigure}[t]{0.44\textwidth}
\includegraphics[width=\textwidth]{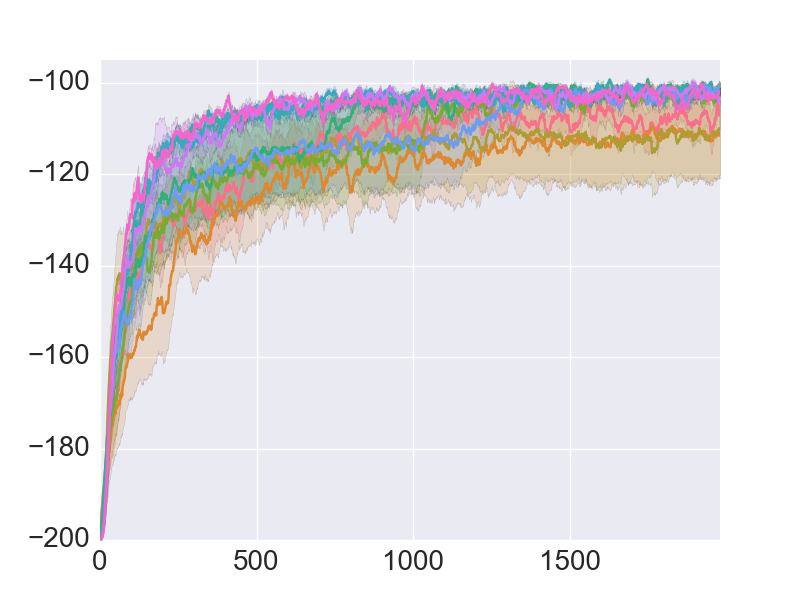}
\caption{$\mu=2^{-12}$}
\end{subfigure}
\hfill
\begin{subfigure}[t]{0.44\textwidth}
\includegraphics[width=\textwidth]{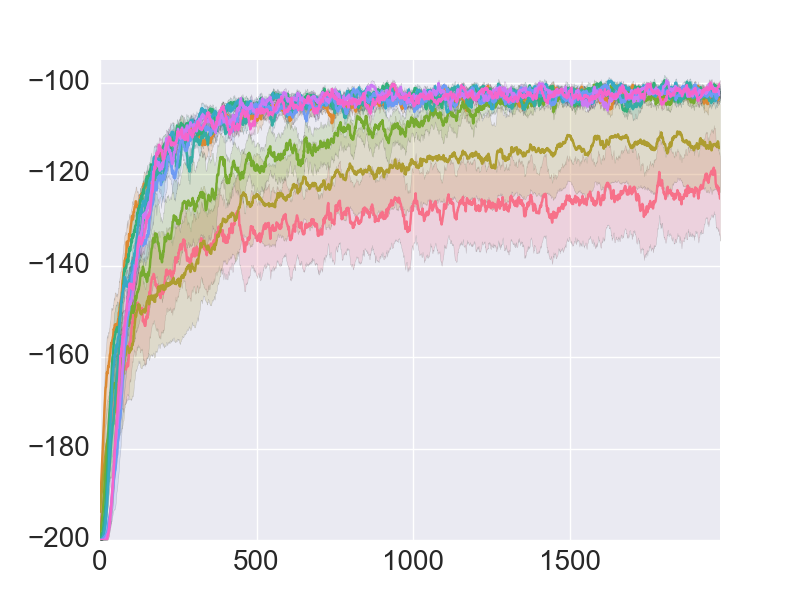}
\caption{$\mu=2^{-13}$}
\end{subfigure}
\linebreak
\begin{subfigure}[t]{0.44\textwidth}
\includegraphics[width=\textwidth]{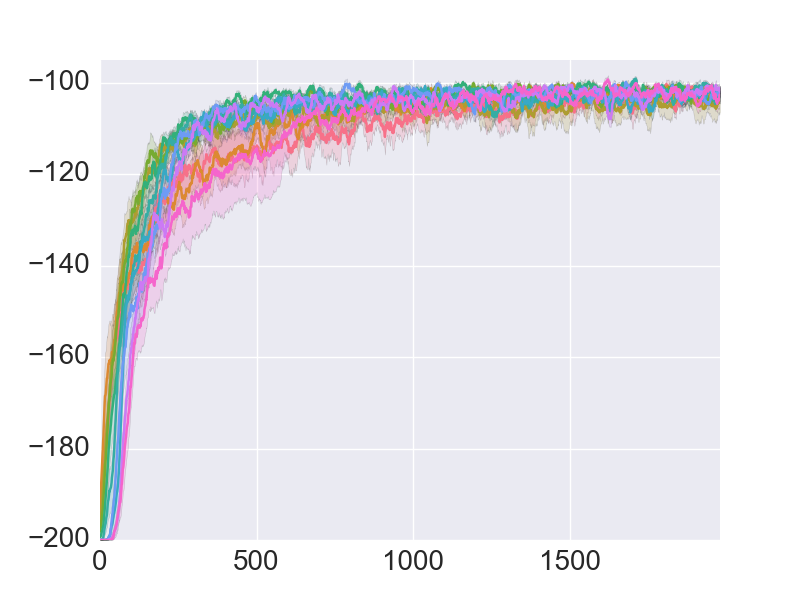}
\caption{$\mu=2^{-14}$}
\end{subfigure}
\hfill
\begin{subfigure}[t]{0.44\textwidth}
\includegraphics[width=\textwidth]{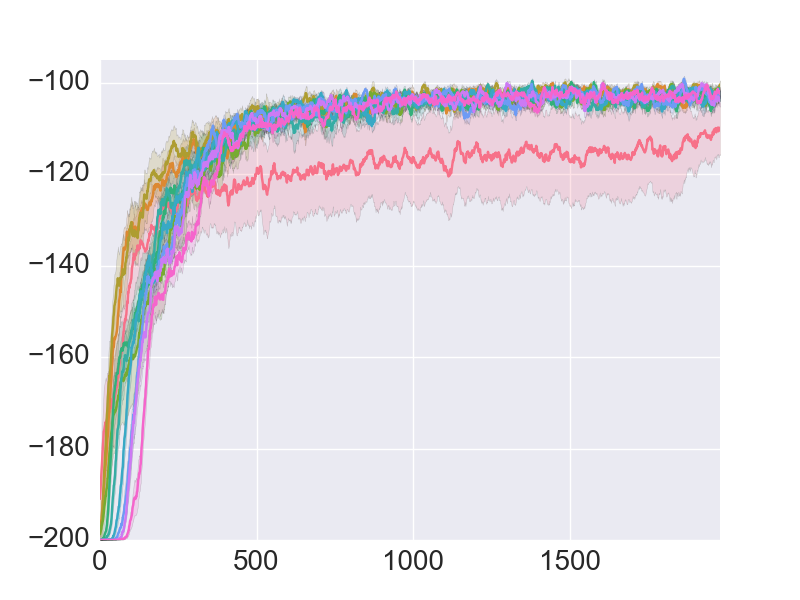}
\caption{$\mu=2^{-15}$}
\end{subfigure}
\caption{Return on mountain-car for a variety of initial $\alpha$ values with NOSID with a variety of $\mu$ values. Each curve shows the average of 10 repeats and is smoothed by taking the average of the last 20 episodes. NOSID behaves quite differently from Normalized Scalar Metatrace, likely owing to the difference in normalization technique applied.}
\label{NOSID}
\end{figure}

We performed this same experiment for SID and NOSID \cite{dabney}, modified for \textit{AC}$(\lambda)$ rather than \textit{SARSA}$(\lambda)$. These results are presented in Figures \ref{SID} and \ref{NOSID}. We found that the unnormalized variants of the two algorithms behaved very similarly on this problem, while the normalized variants showed differences. This is likely due to differences in the normalization procedures. Our normalization procedure is based on that found in \cite{autostep}. The one used by NOSID is based on that found in \cite{NOSID}. The normalization of NOSID seems to make the method more robust to initial $\alpha$ values that are too high, as well as maintaining performance across a wider range of $\mu$ values. On the other hand, it does not shift the useful range of $\mu$ values up as much\footnote{NOSID becomes unstable around $2^{-8}$}, again potentially indicating more variation in optimal $\mu$ value across problems. Additionally, the normalization procedure of NOSID makes use of a hard maximum rather than a running maximum in normalizing the $\beta$ update, which is not well suited for non-stationary state representations where the appropriate normalization may significantly change over time. While this problem was insufficient to demonstrate a difference between SID and Metatrace, we conjecture that in less sparse settings where each weight is updated more frequently, the importance of incorporating the correct time index would be more apparent. Our ALE experiments certainly fit this description, however we do not directly compare with the update rule of SID in the ALE domain. For now, we note that \cite{dabney} focuses on the scalar $\alpha$ case and does not extend to vector valued $\alpha$. Comparing these approaches more thoroughly, both theoretically and empirically, is left for future work.

Vector Metatrace is much less effective on this problem (results are omitted). This can be understood as follows: the tile-coding representation is sparse, thus at any given time-step very few features are active. We get very few training examples for each $\alpha_i$ value when vector step-sizes are used. Using a scalar step-size generalizes over all weights to learn the best overall step-size and is far more useful. On the other hand, one would expect vector step-size tuning to be more useful with dense representations or when there is good reason to believe different features require different learning rates. An example of the later is when certain features are non-stationary or very noisy, as we will demonstrate next.

\subsection{Drifting Mountain Car}
Here we extend the mountain car domain, adding noise and non-stationarity to the state representation. This is intended to provide a proxy for the issues inherent in representation learning (e.g., using a NN function approximator), where the features are constantly changing and some are more useful than others. Motivated by the noisy, non-stationary experiment from \cite{idbd}, we create a version of mountain car that has similar properties. We use the same $1600$ tiling features as in the mountain car case but at each time-step, each feature has a chance to randomly flip from indicating activation with $1$ to $-1$ and vice-versa. We use a uniform flipping probability per time step across all features that we refer to as the drift rate. Additionally we add $32$ noisy features which are 1 or 0 with probability $0.5$ for every time-step. In expectation, $16$ will be active on a given time-step. With $16$ tilings, $16$ informative features will also be active, thus in expectation half of the active features will be informative.

Due to non-stationarity, an arbitrarily small $\alpha$ is not asymptotically optimal, being unable to track the changing features. Due to noise, a scalar $\alpha$ will be suboptimal. Nonzero $\alpha_i$ values for noisy features lead to weight updates when that feature is active in a particular state; this random fluctuation adds noise to the learning process. This can be avoided by annealing associated $\alpha_i$s to zero.

\begin{figure}[!t]
\begin{subfigure}[t]{0.48\textwidth}
\includegraphics[width=\textwidth]{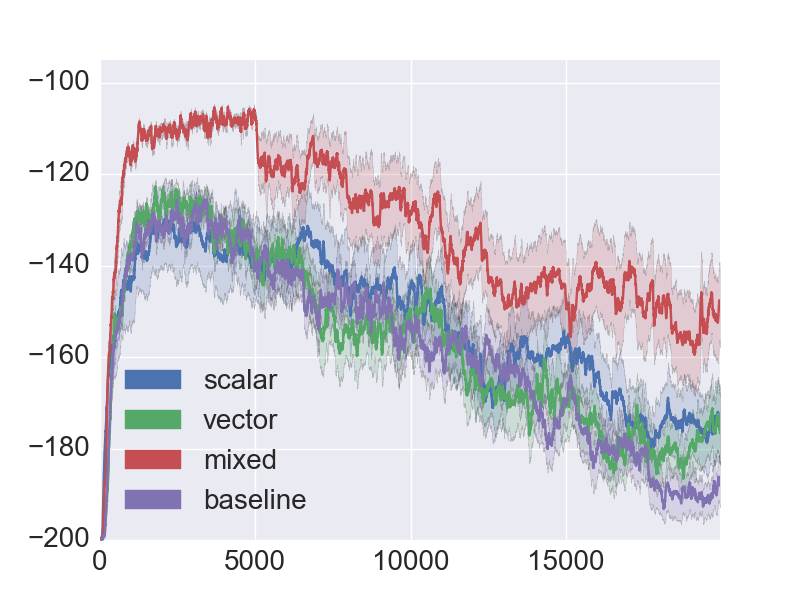}
\caption{drift rate$=4\times 10^{-6}$}
\end{subfigure}
\hfill
\begin{subfigure}[t]{0.48\textwidth}
\includegraphics[width=\textwidth]{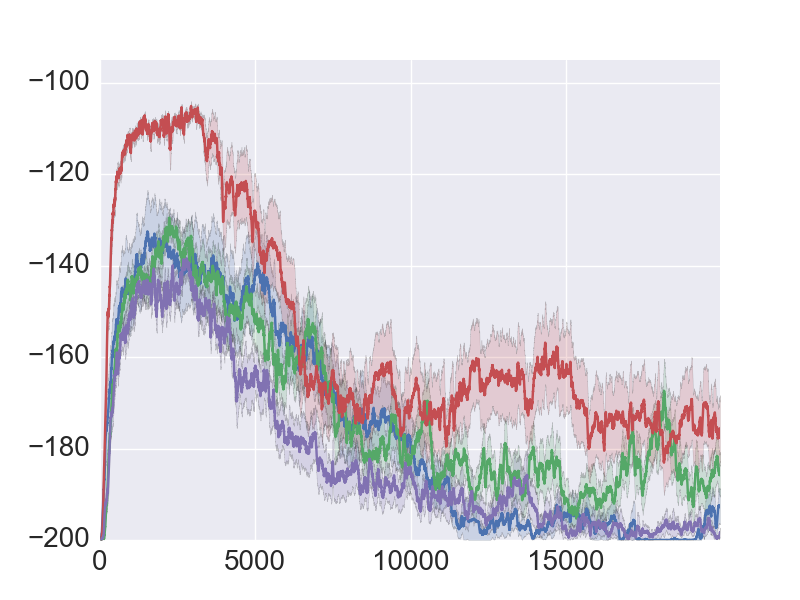}
\caption{drift rate$=6\times 10^{-6}$}
\end{subfigure}
\linebreak
\begin{subfigure}[t]{0.48\textwidth}
\includegraphics[width=\textwidth]{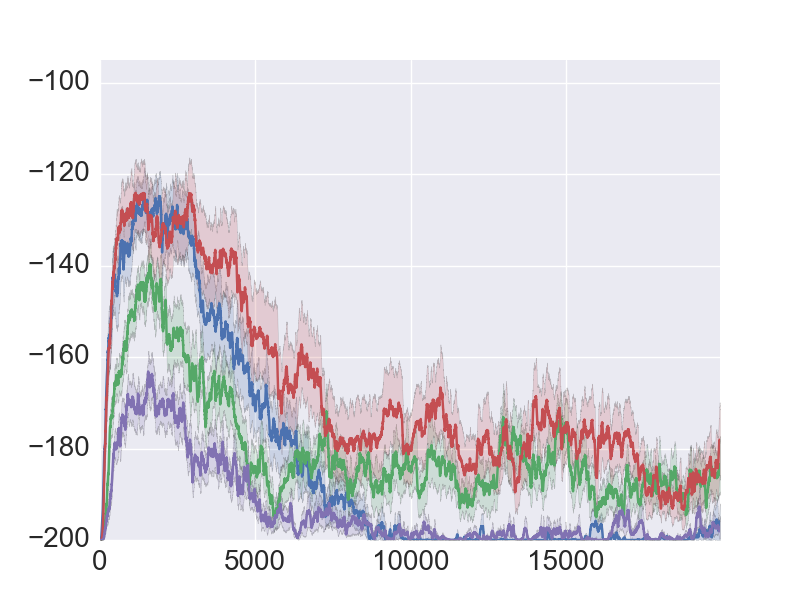}
\caption{drift rate$=8\times 10^{-6}$}
\end{subfigure}
\hfill
\begin{subfigure}[t]{0.48\textwidth}
\includegraphics[width=\textwidth]{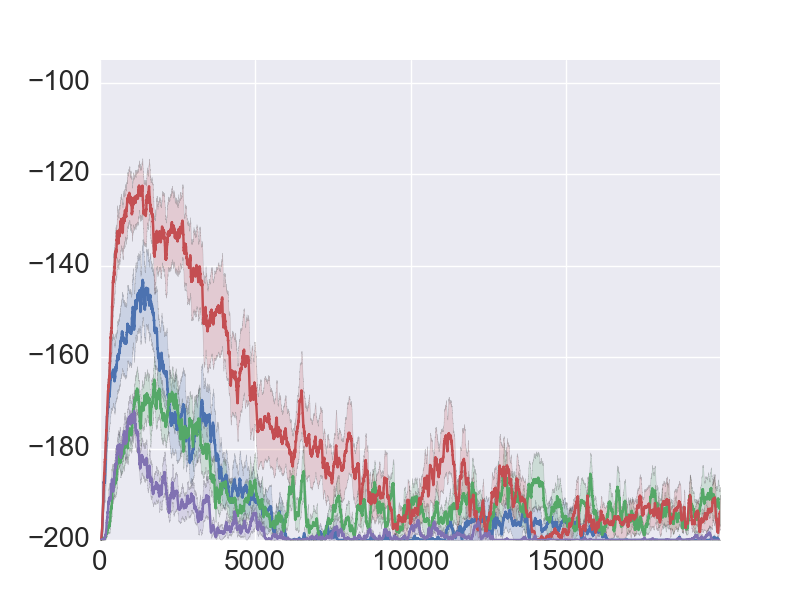}
\caption{drift rate$=1\times 10^{-5}$}
\end{subfigure}
\caption{Return vs. training episodes on drifting mountain car for a fixed initial alpha value of $2^{-10}$ with best $\mu$ value for each tuning method based on average return over the final $100$ episodes. Each curve shows the average of 20 repeats, smoothed over 40 episodes. While all tuning methods improve on the baseline to some degree, mixed Metatrace is generally best.}
\label{drift_best}
\end{figure}

Figure \ref{drift_best} shows the best $\mu$ value tested for several drift rate values for scalar, vector, and mixed Metatrace methods along with a baseline with no tuning. All methods learn quickly initially before many features have flipped from their initial value. Once many features have flipped we see the impact of the various tuning methods. Mixed Metatrace performs the best in the long run so there is indeed some benefit to individually tuning $\alpha$s on this problem. Scalar Metatrace is able to accelerate early learning but at the higher drift values eventually drops off as it is unable to isolate the informative features from the noise. Vector Metatrace tends to under-perform scalar near the beginning but eventually surpasses it as it is eventually able to isolate the informative features from the noise. 
% \FloatBarrier

\begin{figure}[!t]
\begin{subfigure}[t]{0.44\textwidth}
\vskip 0pt
\includegraphics[width=\textwidth]{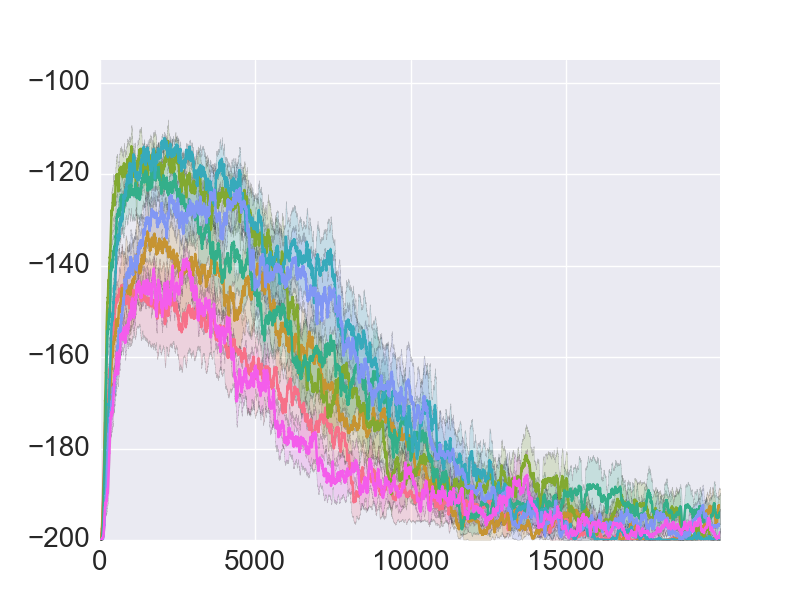}
\caption{Scalar Metatrace $\mu$ sweep.}
\end{subfigure}
\begin{subfigure}[t]
{0.12\textwidth}
\vskip 10pt
\includegraphics[width=\textwidth]{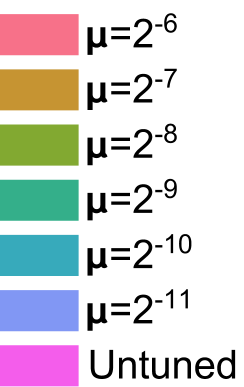}
\end{subfigure}
\begin{subfigure}[t]{0.44\textwidth}
\vskip 0pt
\includegraphics[width=\textwidth]{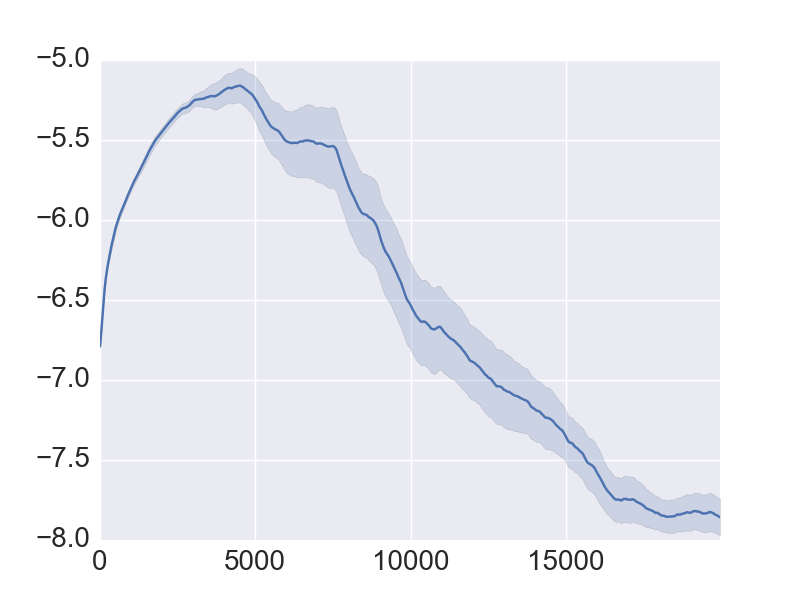}
\caption{Scalar Metatrace $\beta$ evolution.}
\end{subfigure}
\linebreak
\begin{subfigure}[t]{0.44\textwidth}
\includegraphics[width=\textwidth]{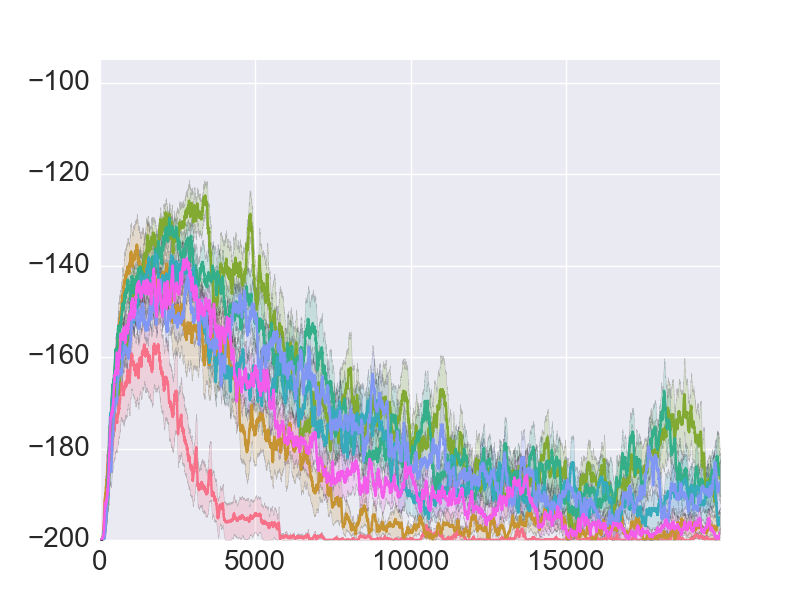}
\caption{Vector Metatrace $\mu$ sweep.}
\end{subfigure}
\hfill
\begin{subfigure}[t]{0.44\textwidth}
\includegraphics[width=\textwidth]{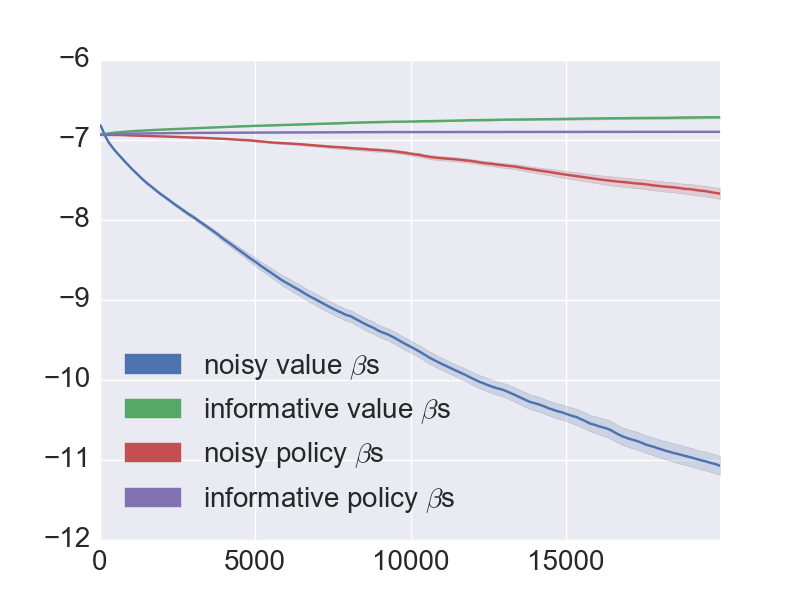}
\caption{Vector Metatrace $\beta$ evolution.}
\end{subfigure}
\linebreak
\begin{subfigure}[t]{0.44\textwidth}
\includegraphics[width=\textwidth]{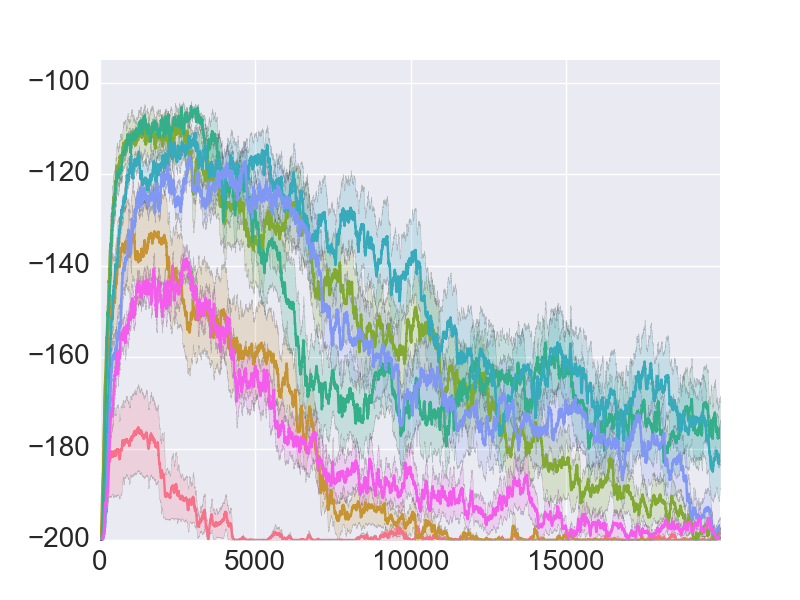}
\caption{Mixed Metatrace $\mu$ sweep.}
\end{subfigure}
\hfill
\begin{subfigure}[t]{0.44\textwidth}
\includegraphics[width=\textwidth]{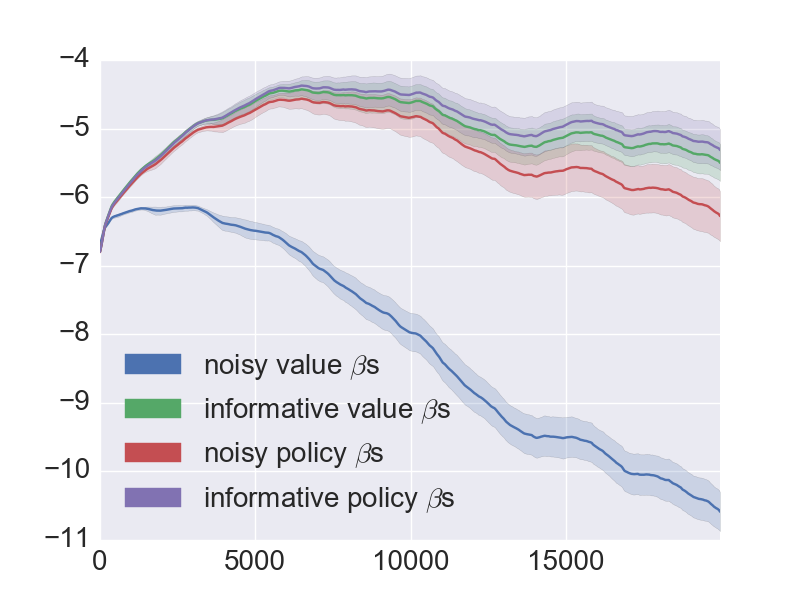}
\caption{Mixed Metatrace $\beta$ evolution.}
\end{subfigure}
\linebreak
\caption{(a, c, e) Return vs. training episodes on drifting mountain car for a fixed initial alpha value of $2^{-10}$ with various $\mu$ value and drift fixed to $6\times 10^{-6}$. (b, d, f) Evolution of average $\beta$ values for various weights on drifting mountain car for different tuning methods for initial $\alpha=2^{-10}$, $\mu=2^{-10}$, drift$=6\times 10^{-6}$. Each curve shows the average of 20 repeats, smoothed over 40 episodes.}
\label{drift_all}
\end{figure}

Figure \ref{drift_all} a, c and e show all $\mu$ values tested for each method for one drift rate value to indicate the sensitivity of each method to $\mu$. The general pattern is similar for a broad range of $\mu$ values. Figure \ref{drift_all} b, d and f show how the average $\beta$ values for different weights evolve over time for the various methods. Both vector and mixed metatrace show much stronger separation between the noisy and informative features for the value function than for the policy. One possible explanation for this is that small errors in the value function have far more impact on optimizing the objective $J_\lambda^\beta$ in mountain car than small imperfections in the policy. Learning a good policy requires a fine-grain ability to distinguish the value of similar states, as individual actions will have a relatively minor impact on the car. On the other hand, errors in the value function in one state have a large negative impact on both the value learning of other states and the ability to learn a good policy. We expect this outcome would vary across problems.

\subsection{Arcade Learning Environment}
%The weights of both output layers were initialized to zero which means the initial policy is uniformly random, and the initial value function is uniformly zero. This maximizes initial exploration, as well as avoiding backing up random initialization values in the TD update.
Here we describe our experiments with the 5 original training set games of the ALE (\textit{asterix}, \textit{beam rider}, \textit{freeway}, \textit{seaquest}, \textit{space invaders}). We use the nondeterministic ALE version with repeat\_action\_probability = 0.25, as endorsed by \cite{ALE}. We use a convolutional architecture similar to that used in the original DQN paper \cite{DQNOG}. Input was 84x84x4, downsampled, gray-scale versions of the last 4 observed frames, normalized such that each input value is between 0 and 1. We use frameskipping such that only every $4^{th}$ frame is observed. We use 2 convolutional layers with 16 8x8 stride 4, and 32 4x4 stride 2 filters, followed by a dense layer of 256 units. Following \cite{SiLU}, activations were dSiLU in the fully connected layer and SiLU in the convolutional layers. Output consists of a linear state value function and softmax policy. We fix $\gamma=0.99$ and $\lambda=0.8$ and use entropy regularization with $\psi=0.01$. For Metatrace we also fix $\mu=0.001$, a value that (based on our mountain car experiments) seems to be reasonable. We run all experiments up to 12.5 million observed frames\footnote{Equivalent to 50 million emulator frames when frame skipping is accounted for.}.

We first performed a broad sweep of $\alpha$ values with only one repeat each for each of the 3 tuning methods, and an untuned baseline on \textit{seaquest} to get a sense of how the different methods perform in this domain. While it is difficult to draw conclusions from single runs, the general trend we observed is that the scalar tuning method helped more initial $\alpha$ values break out of the first performance plateau and made performance of different initial $\alpha$ values closer across the board. Scalar tuning however did not seem to improving learning of the strongest initial $\alpha$s. Vector and mixed tuning, on the other hand, seemed to improve performance of near optimal $\alpha$s as well as reducing the discrepancy between them, but did not seem to help with enabling weaker initial $\alpha$ values to break out of the first performance plateau. Note also, likely owing in part to the normalization, with each tuning method the highest $\alpha$ tested is able to improve to some degree, while it essentially does not improve at all without tuning.
\begin{figure}[!t]
\begin{subfigure}[t]{0.44\textwidth}
\vskip 0pt
\includegraphics[width=\textwidth]{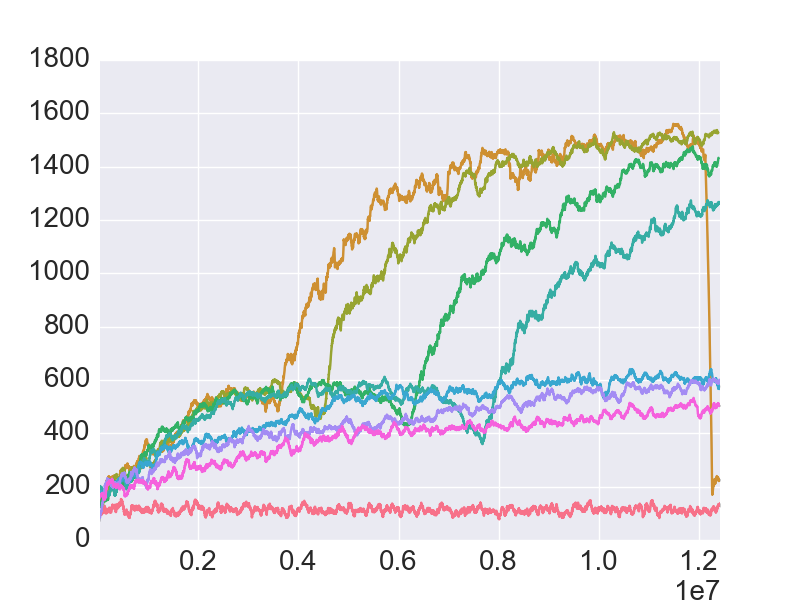}
\caption{No tuning.}
\end{subfigure}
\begin{subfigure}[t]{0.12\textwidth}
\vskip 10pt
\includegraphics[width=\textwidth]{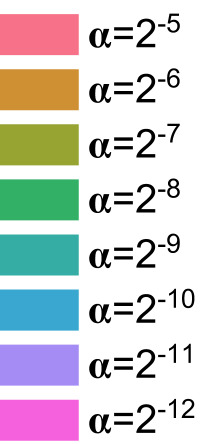}
\end{subfigure}
\begin{subfigure}[t]{0.44\textwidth}
\vskip 0pt
\includegraphics[width=\textwidth]{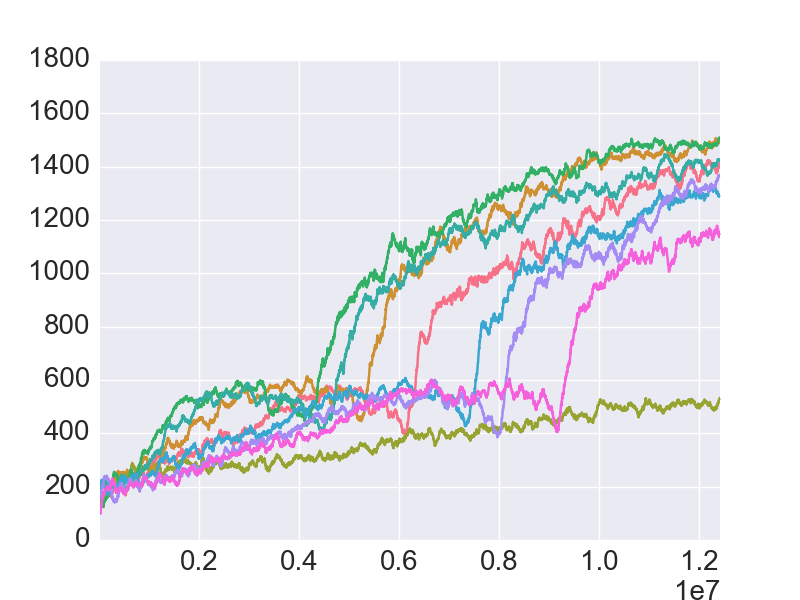}
\caption{Scalar Metatrace.}
\end{subfigure}
\linebreak
\begin{subfigure}[t]{0.44\textwidth}
\includegraphics[width=\textwidth]{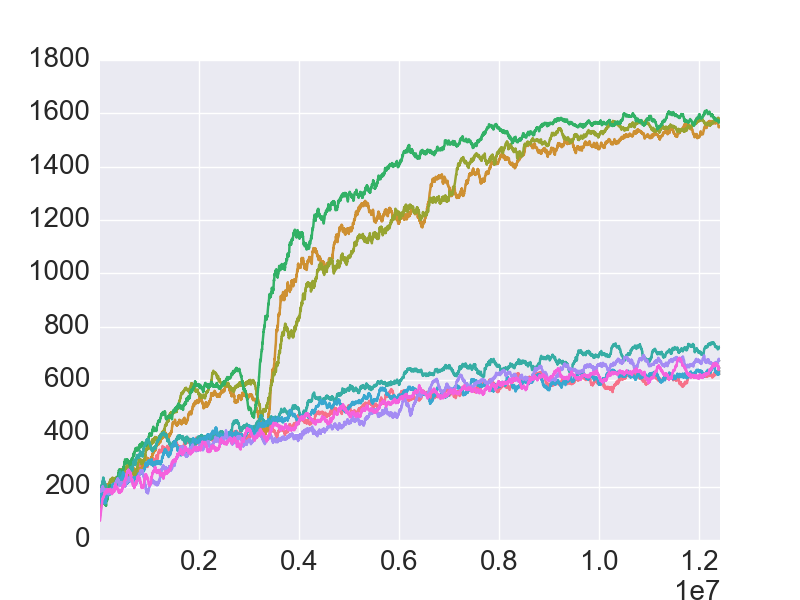}
\caption{Mixed Metatrace.}
\end{subfigure}
\hfill
\begin{subfigure}[t]{0.44\textwidth}
\includegraphics[width=\textwidth]{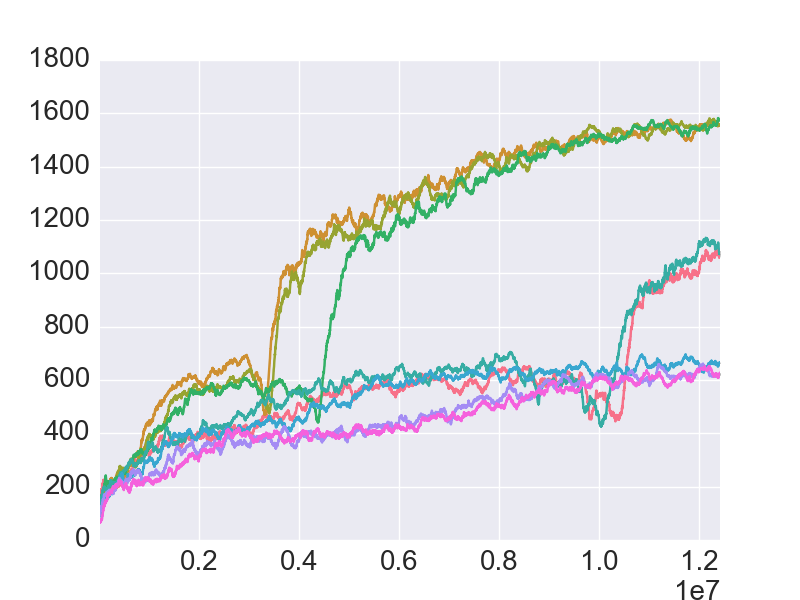}
\caption{Vector Metatrace.}
\end{subfigure}
\caption{Return vs. learning steps for different step-size tuning methods across initial $\alpha$ values for seaquest. In all cases normalization was enabled and $\mu$ fixed to $0.001$.}
\label{ALE_initial}
\end{figure}

After this initial sweep we performed more thorough experiments on the 5 original training set games using mixed Metatrace with the 3 best $\alpha$ values found in our initial sweep. Additionally, we test a baseline with no tuning with the same $\alpha$ values. We ran 5 random seeds with each setting. Figure \ref{ALE_sweep} shows the results of these experiments.
%For reference we note that the average number of episodes completed by the baseline agents in the 12.5 million training steps were as follows: \textit{asterix}: 31833, beam rider: 8993, freeway: 6101, \textit{seaquest}: 11637, space invaders: 18249. Note that \cite{SiLU}, which uses \textit{SARSA}$(\lambda)$ to train a deep neural networks online for the ALE, trains their agents for 200,000 episodes on each game hence on the two games where our experiments overlap (\textit{asterix} and \textit{beamrider}) our results represent far fewer training frames. 
% \FloatBarrier

Metatrace significantly decreases the sensitivity to the initial choice of $\alpha$. In many cases, metatrace also improved final performance while accelerating learning. In space invaders, all 3 $\alpha$ values tested outperformed the best $\alpha$ with no tuning, and in \textit{seaquest} 2 of the 3 did. In \textit{asterix}, the final performance of all initial $\alpha$s with Metatrace was similar to the best untuned alpha value but with faster initial learning. In beam rider, however, we see that using no tuning results in faster learning, especially for a well tuned initial $\alpha$. We hypothesize that this result can be explained by the high $\alpha$ sensitivity and slow initial learning speed in \textit{beam rider}. There is little signal for the $\alpha$ tuning to learn from early on, so the $\alpha$ values just drift due to noise and the long-term impact is never seen. In future work it would be interesting to look at what can be done to make Metatrace robust to this issue. In \textit{freeway}, no progress occurred either with or without tuning.
\begin{figure}
\begin{subfigure}[t]{0.44\textwidth}
\includegraphics[width=\textwidth]{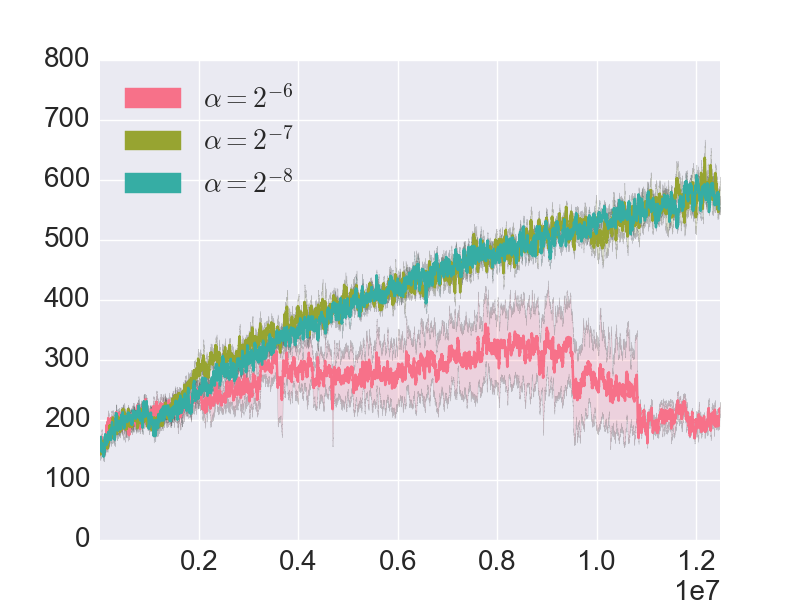}
\caption{No tuning on \textit{space invaders}.}
\end{subfigure}
\hfill
\begin{subfigure}[t]{0.44\textwidth}
\includegraphics[width=\textwidth]{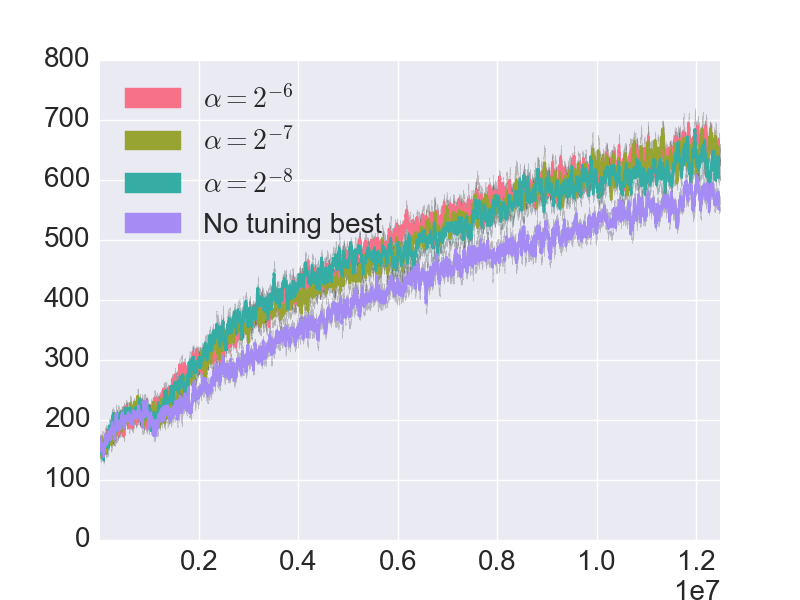}
\caption{Metatrace on \textit{space invaders}.}
\end{subfigure}
\linebreak
\begin{subfigure}[t]{0.44\textwidth}
\includegraphics[width=\textwidth]{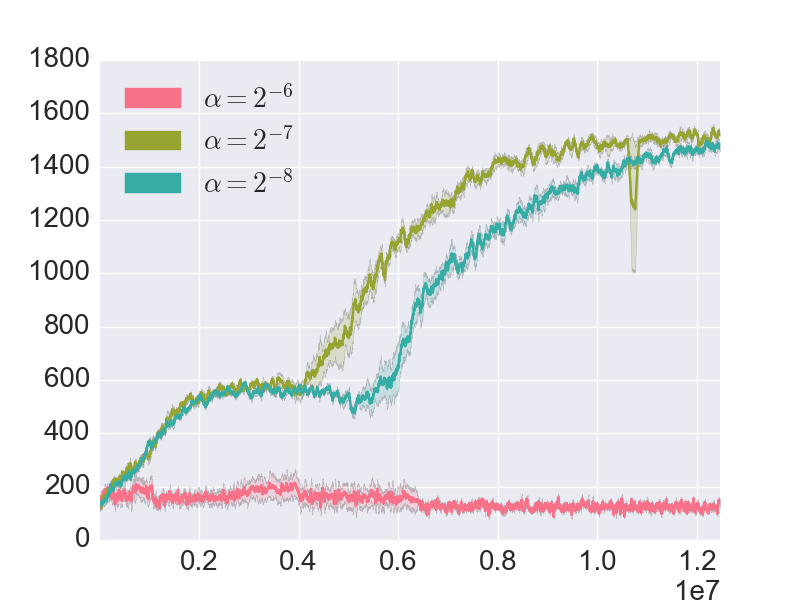}
\caption{No tuning on \textit{seaquest}.}
\end{subfigure}
\hfill
\begin{subfigure}[t]{0.44\textwidth}
\includegraphics[width=\textwidth]{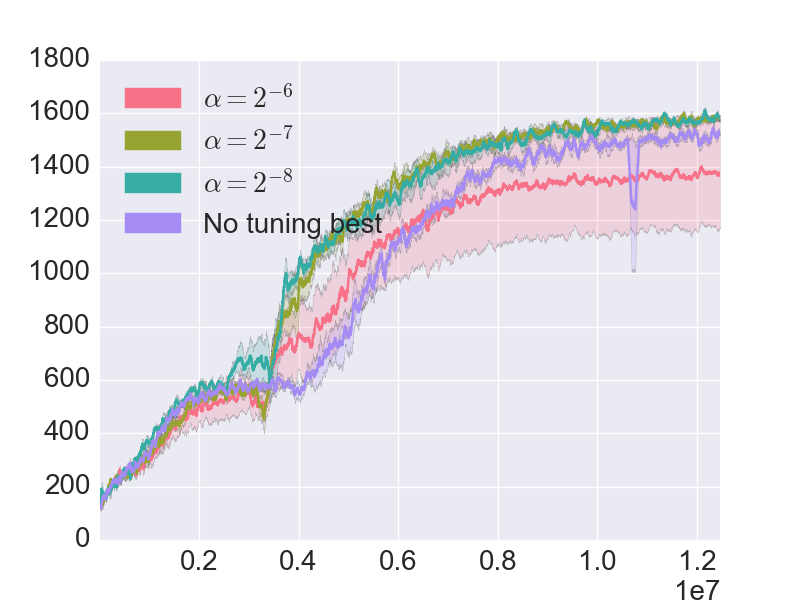}
\caption{Metatrace with on \textit{seaquest}.}
\end{subfigure}
\linebreak
\begin{subfigure}[t]{0.44\textwidth}
\includegraphics[width=\textwidth]{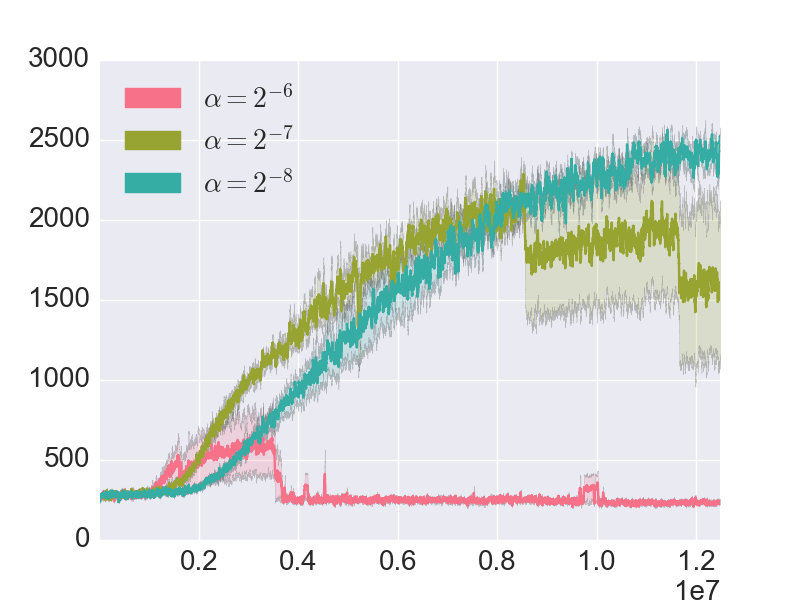}
\caption{No tuning on \textit{asterix}.}
\end{subfigure}
\hfill
\begin{subfigure}[t]{0.44\textwidth}
\includegraphics[width=\textwidth]{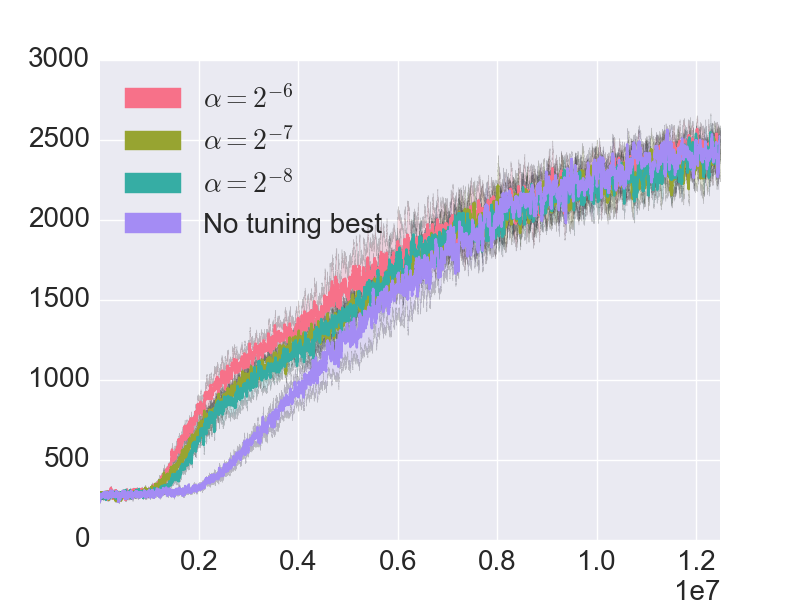}
\caption{Metatrace on \textit{asterix}.}
\end{subfigure}
\end{figure}
\begin{figure}
\ContinuedFloat
\begin{subfigure}[t]{0.44\textwidth}
\includegraphics[width=\textwidth]{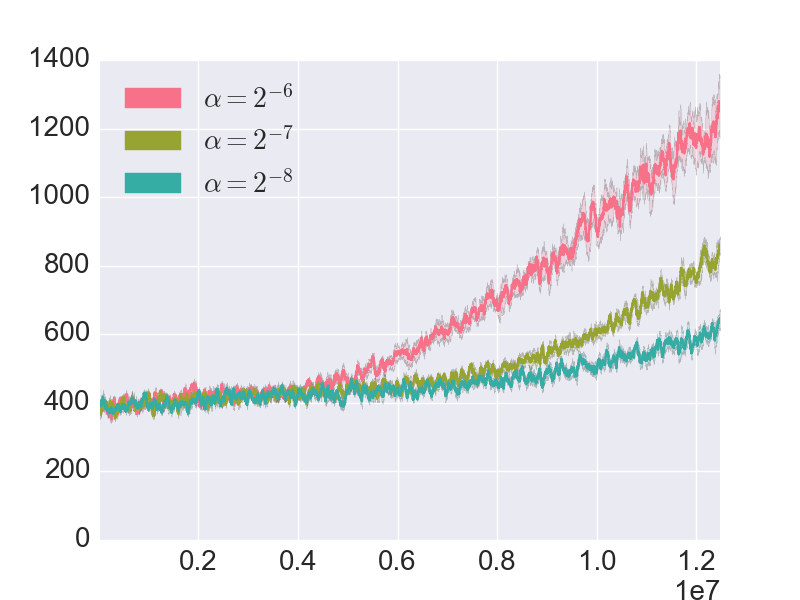}
\caption{No tuning on \textit{beam rider}.}
\end{subfigure}
\hfill
\begin{subfigure}[t]{0.44\textwidth}
\includegraphics[width=\textwidth]{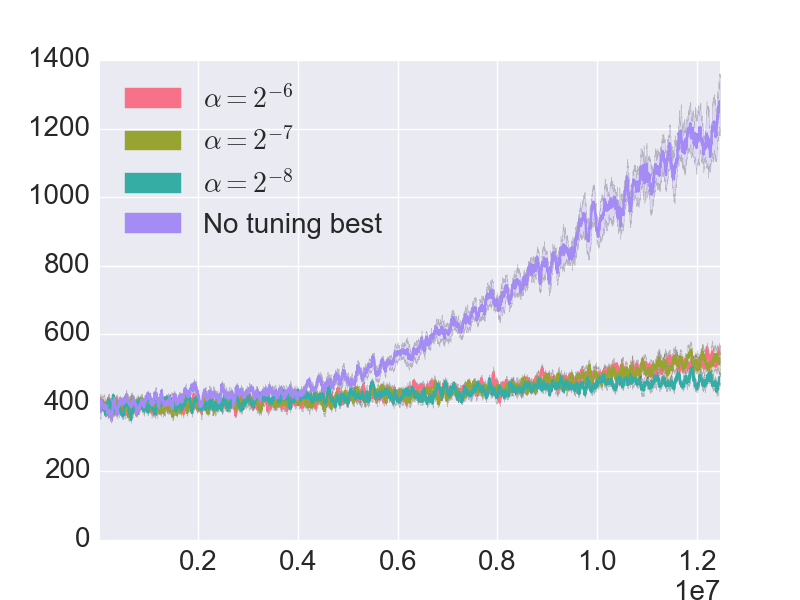}
\caption{Metatrace on \textit{beam rider}.}
\end{subfigure}
\linebreak
\begin{subfigure}[t]{0.44\textwidth}
\includegraphics[width=\textwidth]{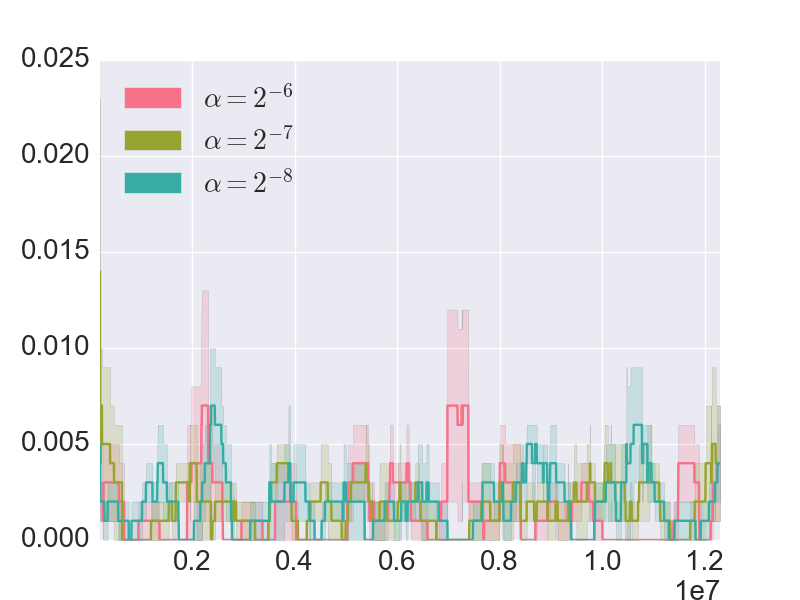}
\caption{No tuning on \textit{freeway}.}
\end{subfigure}
\hfill
\begin{subfigure}[t]{0.44\textwidth}
\includegraphics[width=\textwidth]{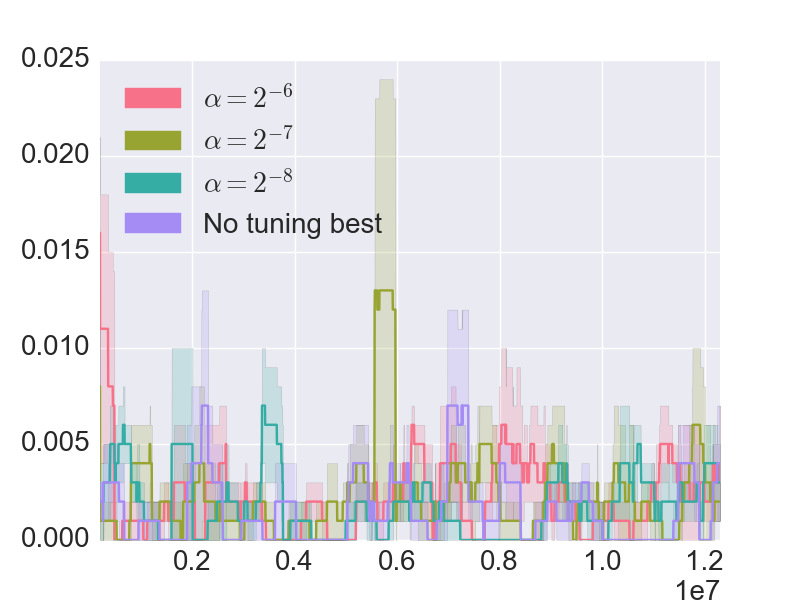}
\caption{Metatrace on \textit{freeway}.}
\end{subfigure}
\caption{Return vs. learning steps for ALE games. Each curve is the average of 5 repeats and is smoothed by taking a running average over the most recent 40 episodes. The meta-step-size parameter $\mu$ was fixed to $0.001$ for each run. For ease of comparison, the plots for Metatrace also include the best tested constant $\alpha$ value in terms of average return over the last 100 training episodes.}
\label{ALE_sweep}
\end{figure}

\section{Conclusion}
We introduce Metatrace, a novel set of algorithms based on meta-gradient descent, which performs step-size tuning for \textit{AC}$(\lambda)$. We demonstrate that Scalar Metatrace improves robustness to initial step-size choice in a standard RL domain, while Mixed Metatrace facilitates learning in an RL problem with non-stationary state representation. The latter result extends results of \cite{idbd} and \cite{autostep} from the SL case. Reasoning that such non-stationarity in the state representation is an inherent feature of NN function approximation, we also test the method for training a neural network online for several games in the ALE. Here we find that in three of the four games where the baseline was able to learn, Metatrace allows a range of initial step-sizes to learn faster and achieve similar or better performance compared to the best fixed choice of $\alpha$.

In future work we would like to investigate what can be done to make Metatrace robust to the negative example we observed in the ALE. One thing that may help here is a more thorough analysis of the nonlinear case to see what can be done to better account for the higher order effects of the step-size updates on the weights and eligibility traces, without compromising the computational efficiency necessary to run the algorithm on-line. We are also interested in applying a similar meta-gradient descent procedure to other RL hyperparameters, for example the bootstrapping parameter $\lambda$, or the entropy regularization parameter $\psi$. More broadly we would like to be able to abstract the ideas behind online meta-gradient descent to the point where one could apply it automatically to the hyperparameters of an arbitrary online RL algorithm.

\section*{Acknowledgements}
The authors would like to acknowledge Pablo Hernandez-Leal, Alex Kearney and Tian Tian for useful conversation and feedback.

\bibliographystyle{unstr}

\end{document}